\documentclass[10pt,twocolumn,letterpaper]{article}

\usepackage[pagenumbers]{cvpr} %

\usepackage{graphicx}
\usepackage{amsmath}
\usepackage{amssymb}
\usepackage{multirow}
\usepackage{booktabs}
\usepackage[accsupp]{axessibility}
\usepackage[ruled,vlined]{algorithm2e}
\usepackage[dvipsnames]{xcolor}
\definecolor{commentcolor}{RGB}{59,116,116}   %
\newcommand{\PyComment}[1]{\ttfamily\textcolor{commentcolor}{\# #1}}  %
\newcommand{\PyCode}[1]{\ttfamily\textcolor{black}{#1}} %
\usepackage[pagebackref,breaklinks,colorlinks]{hyperref}

\usepackage[capitalize]{cleveref}
\crefname{section}{Sec.}{Secs.}
\Crefname{section}{Section}{Sections}
\Crefname{table}{Table}{Tables}
\crefname{table}{Tab.}{Tabs.}

\def\cvprPaperID{4147} %
\def\confName{CVPR}
\def\confYear{2023}

\usepackage{color}
\usepackage{ifthen}

\definecolor{zhiqiu_color}{rgb}{0,0.5,1}
\definecolor{deva_color}{rgb}{0.2,.64,0}
\definecolor{deepak_color}{rgb}{1,0,1}
\definecolor{sam_color}{rgb}{0.3,0.5,0.5}
\definecolor{amelia_color}{rgb}{0.6,.64,0}

\newif\ifsubmit
\submitfalse
\ifsubmit
    \newcommand{\zhiqiu}[1]{}
    \newcommand{\deva}[1]{}
    \newcommand{\deepak}[1]{}
    \newcommand{\sam}[1]{}
    \newcommand{\amelia}[1]{}
\else
    \newcommand{\zhiqiu}[1]{\textsf{\textcolor{zhiqiu_color}{[{\bf Zhiqiu}: #1]}}}
    \newcommand{\deva}[1]{\textsf{\textcolor{deva_color}{[{\bf Deva}: #1]}}}
    \newcommand{\deepak}[1]{\textsf{\textcolor{deepak_color}{[{\bf Deepak}: #1]}}}
    \newcommand{\sam}[1]{\textsf{\textcolor{sam_color}{[{\bf Sam}: #1]}}}
    \newcommand{\amelia}[1]{\textsf{\textcolor{amelia_color}{[{\bf Amelia}: #1]}}}
\fi

\usepackage{amsmath}
\usepackage{bbm}
\DeclareMathOperator*{\argmax}{arg\,max}

\begin{document}
\title{Multimodality Helps Unimodality: \\
Cross-Modal Few-Shot Learning with Multimodal Models}

\author{Zhiqiu Lin* \qquad Samuel Yu* \qquad Zhiyi Kuang \qquad Deepak Pathak \qquad Deva Ramanan\vspace{1mm}\\
Carnegie Mellon University\\
{\tt\small \{zhiqiul,samuelyu,zkuang,dpathak,deva\}@cs.cmu.edu}
}
\maketitle
\def\thefootnote{*}\footnotetext{Equal contribution. Published at CVPR 2023.}
\def\thefootnote{\arabic{footnote}}
\begin{abstract}

    The ability to quickly learn a new task with minimal instruction -- known as few-shot learning -- is a central aspect of intelligent agents. Classical few-shot benchmarks make use of few-shot samples from a single modality, but such samples may not be sufficient to characterize an entire concept class. In contrast, humans use cross-modal information to learn new concepts efficiently. In this work, we demonstrate that one can indeed build a better {\bf visual} dog classifier by {\bf read}ing about dogs and {\bf listen}ing to them bark. To do so, we exploit the fact that recent multimodal foundation models such as CLIP learn cross-modal encoders that map different modalities to the same representation space. Specifically, we propose a simple strategy for {\bf cross-modal adaptation}: we treat examples from different modalities as additional few-shot examples. For example, by simply repurposing class names as an additional training sample, we trivially turn any n-shot learning problem into a (n+1)-shot problem. This allows us to produce SOTA results with embarrassingly-simple linear classifiers. We show that our approach can be combined with existing methods such as prefix tuning, adapters, and classifier ensembling. Finally, to explore other modalities beyond vision and language, we construct the first (to our knowledge) audiovisual few-shot benchmark and use cross-modal training to improve the performance of both image and audio classification. %
    Project site at \href{https://linzhiqiu.github.io/papers/cross_modal/}{link}.
   
\end{abstract}

\section{Introduction}
\label{sec:intro}

\begin{figure}[t!]
    \centering
    \includegraphics[width=0.7\linewidth, clip=true,trim = 0mm 0mm 0mm 0mm]{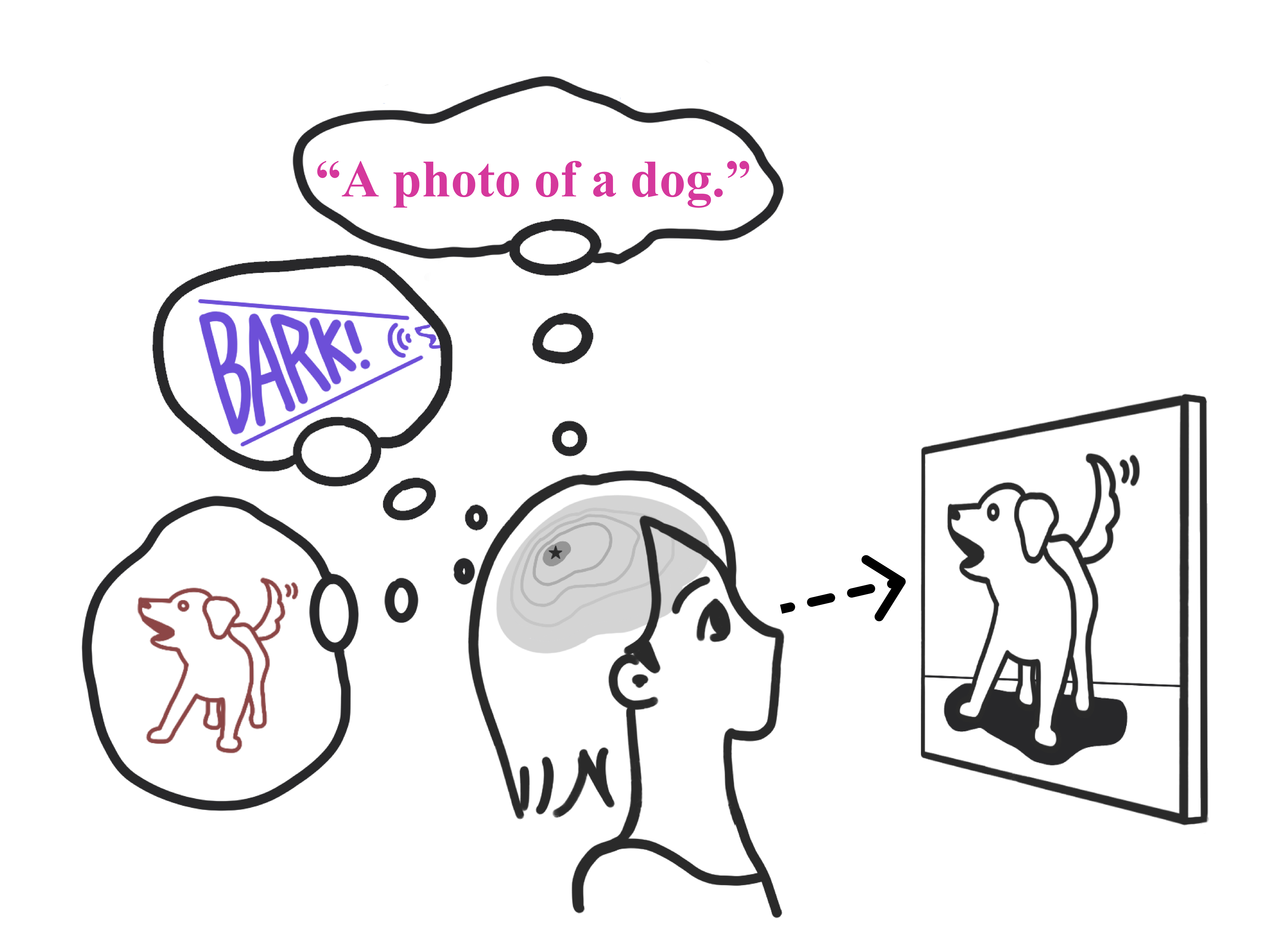}
    \caption{{\bf Human perception is cross-modal.} Our work is loosely inspired by neuroscience studies that suggest that neurons can be triggered from stimuli from different modalities, such as vision, audio, or even language ~\cite{gibson1969principles, meltzoff1979intermodal, Nanay2018-NANMMI}. %
    In this work, we propose to leverage such cross-modality representations to adapt multimodal models (such as CLIP~\cite{radford2021learning} and AudioCLIP~\cite{guzhov2021audioclip}) for few-shot learning with a simple but effective strategy; we learn (non)linear classifiers built on top of few shot examples that span different modalities, including vision, audio, and language (Fig.~\ref{fig:pca_teaser}).} %
  \label{fig:neuro_teaser}
\end{figure}

Learning with minimal instruction is a hallmark of human intelligence~\cite{schmidt2009meaning, wang2018low, snell2017prototypical}, and is often studied under the guise of few-shot learning. In the context of few-shot visual classification~\cite{hariharan2017low, dhillon2019baseline, joachims1999transductive, qi2018low, finn2017model, ravi2016optimization}, a classifier is first pre-trained on a set of base classes to learn a good feature representation and then adapted or finetuned on a small amount of novel class data. However, such few-shot setups often face an inherent ambiguity -- if the training image contains a golden retriever wearing a hat, how does the learner know if the task is to find {\tt dogs}, {\tt golden retrievers}, or even {\tt hats}? On the other hand, humans have little trouble understanding and even generalizing from as few as one example. How so?

We argue that humans make use of multimodal signals and representations (\autoref{fig:neuro_teaser}) when learning concepts. For example, verbal language has been shown to help toddlers better recognize visual objects given just a few examples~\cite{jackendoff1987beyond,smith2005development}. Indeed, there exists ample evidence from neuroscience suggesting that cognitive representations are inherently multimodal. For instance, visual images of a person evoke the same neurons as the textual strings of the person's name~\cite{quiroga2005invariant} and even audio clips of that person talking~\cite{Nanay2018-NANMMI}.
 Even for infants as young as 1-5 months old, there is a strong correspondence between auditory-visual~\cite{kuhl1984intermodal} as well as visual-tactile signals~\cite{meltzoff1979intermodal}. Such \textit{cross-modal} or inter-modal representations are fundamental to the human perceptual-cognitive system, allowing us to understand new concepts even with few examples~\cite{gibson1969principles}. 

\begin{figure}[t!]
    \centering
    \includegraphics[width=\linewidth, clip=true,trim = 0mm 0mm 0mm 0mm]{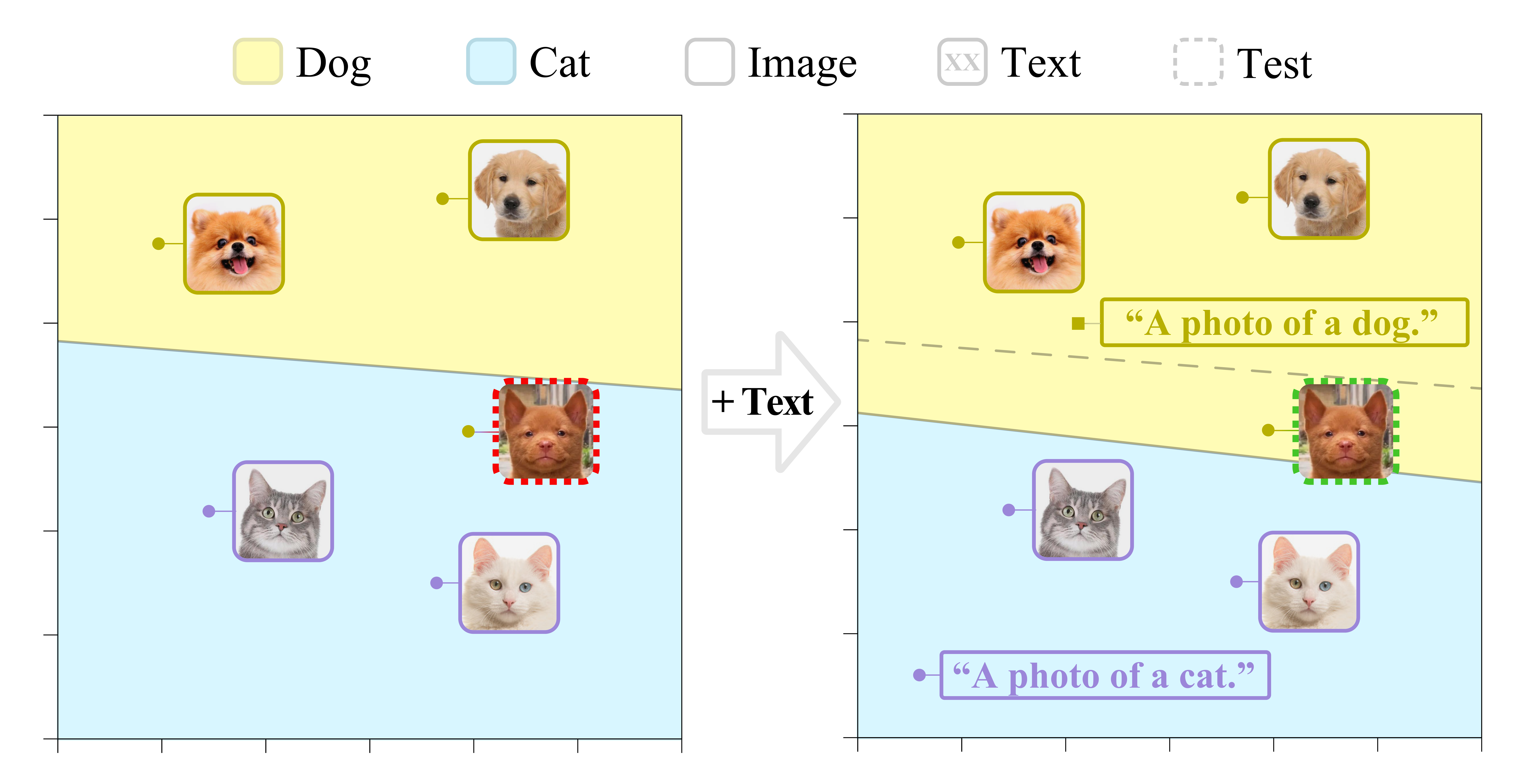}
    \caption{{\bf Adding additional modalities helps few-shot learning}. Adding textual labels to a 2-shot cat-vs-dog classification task leads to better test performance (by turning the problem into a 3-shot cross-modal task!). We visualize cross-modal CLIP~\cite{gao2021clip} features (projection to 2D with principal component analysis) and the resulting classifier learned from them, and observe a large shift in the decision boundary. See \autoref{fig:pcas} for more examples.}
  \label{fig:pca_teaser}
\end{figure}

{\bf Cross-modal adaptation (our approach).}
In this paper, we demonstrate that cross-modal understanding of different modalities (such as image-text or image-audio) can improve the performance of individual modalities. That is, {\em read}ing about dogs and {\em listen}ing to them bark can help build a better {\em visual} classifier for them! To do so, we present a remarkably simple strategy for cross-modal few-shot adaptation: {\em we treat examples from different modalities as additional few-shot examples}. For example, given the ``1-shot" task of learning a {\tt dog} classifier, we treat {\em both} the textual {\tt dog} label and the single visual image as training examples for learning a (visual) dog classifier. Learning is straightforward when using frozen textual and visual encoders, such as CLIP~\cite{radford2021learning}, that map different modalities to the same representational space. In essence, we have converted the ``n-shot" problem to a ``(n+1)-shot" problem (\autoref{fig:pca_teaser})! We demonstrate that this basic strategy produces SOTA results across the board with a simple linear classifier, and can be applied to existing finetuning methods~\cite{zhou2022coop, zhang2021tip, wortsman2022robust} or additional modalities (e.g. audio).

\begin{figure}[t!]
    \centering
    \includegraphics[width=0.9\linewidth, clip=true,trim = 0mm 0mm 0mm 0mm]{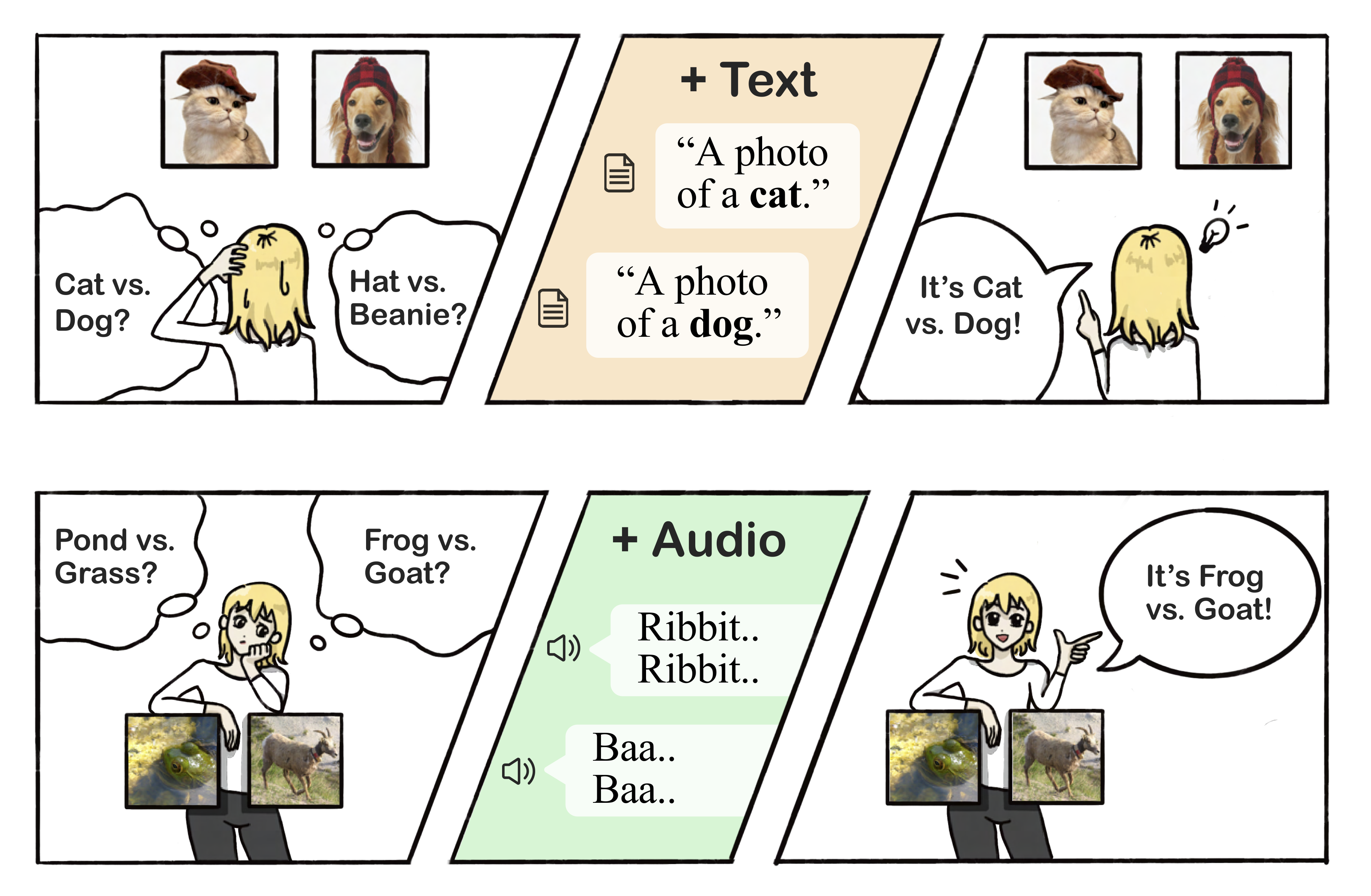}
    \vspace{-3mm}
    \caption{{\bf Cross-modality reduces the ambiguity of few-shot learning.} Classic (uni-modal) few-shot learning is often {\em under}specified. Even for binary classification, when given only a single image per class ({\bf left}), it is unclear whether the target class is the animal, the hat, or the background scene. Adding an extra modality, such as text or audio, helps clarify the problem setup ({\bf right}). Notably, language usually comes ``for free" in classification datasets in the form of a textual label per class.}
    \vspace{-3mm}
  \label{fig:ambiguity}
\end{figure}

{\bf Why does it work?} From one perspective, it may not be surprising that cross-modal adaptation improves accuracy, since it takes advantage of additional training examples that are ``hidden" in the problem definition, e.g. a label name~\cite{xing2019adaptive} or an annotation policy~\cite{mu2019shaping} for each class. However, our experiments demonstrate that multimodal cues are often complementary since they capture different aspects of the underlying concept; a {\tt dog} label paired with a single visual example is often more performant than two images! For example, ~\autoref{fig:ambiguity} demonstrates a one-shot example where the target concept is ambiguous, but becomes clear once we add information from other modalities like language and sound.

{\bf Multimodal adaptation (prior art).} %
In contrast to our cross-modal approach, most prior works simply follow the popular practice of finetuning uni-modal foundation models, such as large vision~\cite{he2020momentum, chen2020simple, he2022masked} or language models~\cite{devlin2018bert, liu2021gpt, brown2020gpt3}.
For example, CoOp~\cite{zhou2022coop} and other prompting methods~\cite{zhu2022prompt, zhou2022cocoop, lu2022prompt} finetune CLIP via prefix tuning to replace hand-engineered prompts such as {\tt "a photo of a \{cls\}"} with learned word tokens. Similarly, inspired by parameter-efficient tuning of language models~\cite{houlsby2019parameter}, adapter-based methods~\cite{gao2021clip, zhang2021tip} finetune CLIP by inserting lightweight multi-layer-perceptrons (MLPs). However, we aim to study the fundamental question of how to finetune {\em multi}-modal (as opposed to {\em uni}-modal) models. A crucial difference between prior art and ours is the use of textual information, as all existing methods~\cite{zhou2022coop, wortsman2022robust, ilharco2022patching, zhang2021tip} repurpose additional text features as {\em classifier weights} instead of {\em training samples}. We demonstrate in this paper that cross-modal adaptation is not only more performant but can also benefit prior uni-modal approaches.

{\bf Problem setup.}
We begin by replicating the existing evaluation protocol of other works~\cite{radford2021learning, zhou2022coop, zhang2021tip} on few-shot adaptation of vision-language models, and report performance on 11 diverse downstream datasets. We produce state-of-the-art accuracy with an embarrassingly simple linear classifier that has access to additional ``hidden"  training examples in the form of textual labels, resulting in a system that is far more lightweight than prior art. Interestingly, we show that existing approaches~\cite{zhou2022coop, zhang2021tip, wortsman2022robust}, despite already repurposing text features as classifier weights, can still benefit from cross-modal learning.
Finally, we extend our work to the audio domain by taking advantage of AudioCLIP~\cite{guzhov2021audioclip} that maps audio to the same frozen CLIP representation space. We construct the first (to our knowledge) {\em cross-modal few-shot learning benchmark with audio} by intersecting ImageNet~\cite{deng2009imagenet} and the ESC-50 audio classification dataset~\cite{piczak2015esc}. We show that cross-modal audiovisual learning helps for both downstream image and audio classification; in summary, one {\em can} train better dog image classifiers by listening to them bark!

\section{Related Works}
\label{sec:related}
{\bf Webly-supervised pre-training.} Learning {\em foundation models}~\cite{bommasani2021opportunities} from large-scale web data is becoming a predominant paradigm in AI. In NLP, models such as BERT~\cite{devlin2018bert} and GPT-3~\cite{brown2020gpt3} are pre-trained on a massive web text corpus with language-modeling objectives and can be transferred to a wide range of downstream tasks, even without explicit supervised finetuning~\cite{tsimpoukelli2021multimodal, liu2021pre}. Self-supervision~\cite{he2020momentum, chen2020simple, caron2021emerging} is also a trending topic in the vision community, and recent methods~\cite{he2022masked, goyal2021self} demonstrate even stronger visual representations than fully-supervised pre-trained ones such as on ImageNet~\cite{deng2009imagenet}. 

{\bf Multimodal foundation models.} Recently, foundation models have shifted towards a multimodal supervision paradigm. For visual representation learning, early works transform web image captions into structured outputs for supervised learning, such as multi-label targets~\cite{joulin2016learning} or visual n-grams~\cite{li2017learning}. More recently, CLIP~\cite{radford2021learning} and ALIGN~\cite{jia2021scaling} propose a simple contrastive-based approach to embed images and captions into the same representation space, and demonstrate impressive ``zero-shot" performance on downstream tasks. Follow-up works enhance multimodal pre-training by incorporating generative-based objectives~\cite{alayrac2022flamingo, li2022blip, yu2022coca}, consistency regularization~\cite{mu2022slip, li2021supervision}, stronger visual priors~\cite{zhai2022lit}, phrase-grounding tasks~\cite{li2021grounded, zhang2022glipv2}, and audiovisual information through videos~\cite{guzhov2021audioclip}. In this work, we focus on adapting CLIP~\cite{radford2021learning} and AudioCLIP~\cite{guzhov2021audioclip} for few-shot classification because contrastive-based multimodal models are stronger classifiers~\cite{alayrac2022flamingo}. Adopting other multimodal models~\cite{alayrac2022flamingo, yu2022coca} or adapting to tasks other than classification~\cite{song2022clip, zhang2022glipv2} can be interesting future directions. 

{\bf Adaptation of foundation models.} 
As multimodal pre-trained models have excelled at classic vision tasks~\cite{zhang2022glipv2, radford2021learning}, there has been surging interest in developing more efficient adaptation methods. However, we observe that most of the trending techniques are built upon successful recipes crafted for uni-modal foundation models. For example, CLIP~\cite{radford2021learning} adopts linear probing~\cite{he2020momentum, zhang2022glipv2, chen2020simple, he2022masked} and full-finetuning~\cite{he2022masked, wang2017growing, zhang2022glipv2, kirkpatrick2017overcoming, girdhar2017attentional, wu2022transferring} when transferring to downstream tasks. Prompt adaptation of CLIP~\cite{radford2021learning, lu2022prompt, zhu2022prompt, xing2022class, zhou2022cocoop} is motivated by the success of prefix-tuning for language models~\cite{liu2021pre, deng2022rlprompt, jiang2020can, prasad2022grips, gao2020making, haviv2021bertese, shin2020autoprompt, schick2020exploiting,schick2020small}. Similarly, CLIP-Adapter~\cite{gao2021clip} and Tip-Adapter~\cite{zhang2021tip} are inspired by parameter-efficient finetuning methods~\cite{houlsby2019parameter, zhang2020side, jia2022visual} that optimize lightweight MLPs while freezing the encoder. Yet, all aforementioned methods including WiSE-FT~\cite{wortsman2022robust} use the other modality, e.g. textual labels, as {\em classifier weights} and still calculate a {\em uni-modal} softmax loss on the few-shot images. We instead show that incorporating other modalities as {\em training samples} is far more effective.

{\bf Few-shot classification.} Prior successful few-shot learning methods leverage meta learning~\cite{finn2017model, ravi2016optimization}, metric learning~\cite{snell2017prototypical, NIPS2016_90e13578, bateni2020improved}, transfer learning~\cite{hariharan2017low, qi2018low}, and transductive learning~\cite{dhillon2019baseline, joachims1999transductive}. These classic algorithms usually assume a large meta-training set for pre-training the network, and then evaluate on multiple episodes of few-shot train (support) and test (query) sets. In this work, we instead follow the new evaluation protocol implemented by recent works on few-shot adaptation with CLIP~\cite{radford2021learning, zhou2022coop, zhang2021tip}: (1) the meta-training phase is replaced with pre-trained CLIP models, and (2) the test sets are the official test splits of each dataset (thus not few-shot). Notably, none of the prior works~\cite{zhang2021tip, zhou2022coop} we compare to in this paper perform optimization with test set samples, and we follow this practice to ensure a fair comparison. We leave semi-supervised~\cite{wang2022debiased} or transductive finetuning~\cite{dhillon2019baseline, huang2022unsupervised} techniques as future work.

{\bf Cross-modal machine learning.} 
Inspired by cross-modal human cognition~\cite{Calvert1997lip, Nanay2018-NANMMI, kosslyn2010brainimagery}, cross-modal learning~\cite{xing2019adaptive, mu2019shaping} is a subfield of multimodal machine learning~\cite{lee2019touching, pahde2021multimodal, pahde2018discriminative, afham2021rich, schwartz2022baby, hong2020x, zhang2021cross, alwassel2020self, li2020unimo, cangea2019xflow, luo2018vitac} that aims to use data from additional modalities to improve a uni-modal task. Cross-modal learning does not require instance-wise alignment; for example, existing algorithms~\cite{xing2019adaptive, mu2019shaping} can benefit from class-level descriptions as opposed to image-level captions. In this work, we propose a lightweight cross-modal learning method by treating data from other modalities as additional training samples. Furthermore, we encourage future works to embrace cross-modal few-shot learning as opposed to the underspecified uni-modal setup (\autoref{fig:ambiguity}).

\section{Cross-Modal Adaptation}
\label{sec:vision-language}
\begin{figure}[t!]
    \centering
    \includegraphics[width=\columnwidth, clip=true,trim = 0mm 0mm 0mm 0mm]{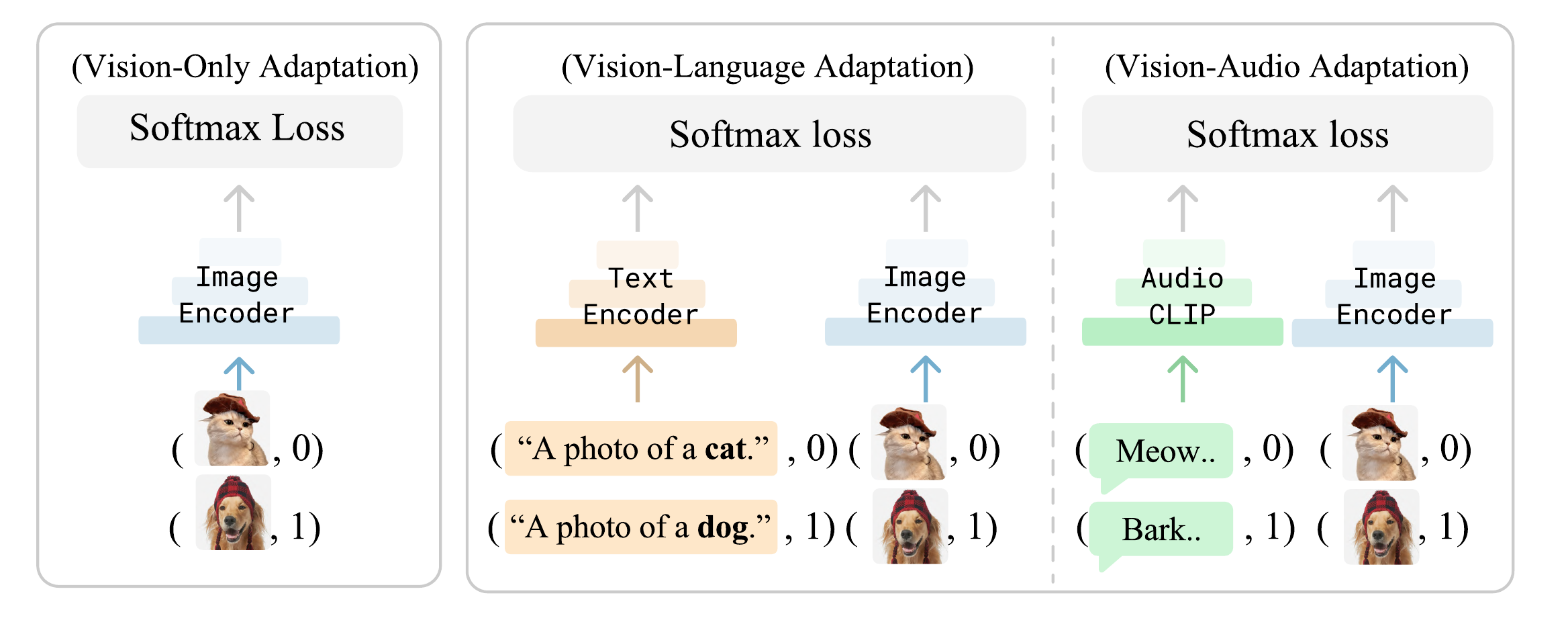}
    \vspace{-8mm}
    \caption{{\bf Uni-modal (left) vs. cross-modal adaptation (right)} for a binary cat-vs-dog classification task. Prior work~\cite{zhou2022coop, zhang2021tip, gao2021clip, wortsman2022robust} optimizes over a loss from a single modality. Cross-modal adaptation makes use of additional training samples from other modalities, exploiting pre-trained encoders that map different modalities to the same representation space. We show that cross-modal learning can also improve prior art and even extends to audio modalities with AudioCLIP~\cite{guzhov2021audioclip}.}
  \label{fig:method}
\end{figure}

In this section, we mathematically formalize our approach to cross-modal few-shot learning. %

{\bf Uni-modal learning.} We begin by reviewing standard uni-modal few-shot classification, which learns a classifier from a small dataset of $(x_i,y_i)$ pairs and pre-trained feature encoder $\phi(\cdot)$: 
\begin{align}
    \label{uni_modal_eq}
    \mathcal{L}_{uni-modal} = \sum_i \mathcal{H}(y_i, \phi(x_i))
\end{align}
\noindent where $\mathcal{H}$ is typically the softmax loss
\begin{align}
    \label{softmax_loss}
    \mathcal{H}(y, f) = -\log \Big( p(y | f) \Big) = - \log \Big( \frac{e^{w_{y} \cdot f}}{\sum_{y'} e^{w_{y'} \cdot f}} \Big).
\end{align}
Our notation separates the feature encoder $\phi$ from the final class weights $w_y$, since the former is typically pre-trained on a massive source dataset and the latter is trained on the few-shot target dataset. However, sometimes the encoder $\phi$ can also be finetuned on the few-shot dataset (as we explore in our experiments). Importantly, both the class weights and the encoder must live in the same $N$-dimensional space in order to compute their inner product:
\begin{align}
    w_y, \phi(\cdot) \in R^N.
\end{align}
Though we focus on classification, class models could be learned via other losses (such as centroid prototypes~\cite{snell2017prototypical}).
 
{\bf Cross-modal learning.} Our extension to multiple modalities is staightforward; we assume each training example is accompanied by a discrete label $m$ denoting its modality: 
\begin{align}
    (x_i,y_i) \rightarrow (x_i,y_i,m_i), \quad x_i \in X_{m_i}, \quad m_i \in M.
\end{align}
For example, one may define the set of modalities to be $M = \{\text{visual},\text{language}\}$ or $\{ \text{visual},\text{audio} \}$ (\autoref{fig:method}). We can then define an associated loss:
\begin{align}
\label{cross_modal_eq}
    \mathcal{L}_{cross-modal} = \sum_i \mathcal{H}(y_i, \phi_{m_i}(x_i)),
\end{align}
\noindent where we crucially assume access to modality-specific encoders $\phi_m$ for $m \in M$. While the individual datapoints $x_i$ may come from different modalities with different dimensions, our formulation requires that the encoders map all modalities to the same fixed-dimensional space.
\begin{align}
    w_y, \phi_m(\cdot) \in R^N.
\end{align}
Note that this requirement is satisfied by many multimodal foundation models such as CLIP~\cite{radford2021learning} and ALIGN~\cite{jia2021scaling} since they map different modalities into the same $N$-dimensional embedding. We provide training pseudocode for vision-language adaptation (\autoref{sec:vision-language}) in \autoref{algo:cross_modal_code} for clarity.

{\bf Inference:} The learned classifier can produce a label prediction for a test example $x$ from {\em any} modality $m \in M$:
\begin{align}
    {\hat y} = \argmax_{y'} w_{y'} \cdot \phi_m(x). \label{eq:inference}
\end{align}
This means we can use the same classifier to classify different test modalities (e.g. images and audio clips). In this paper, we mainly evaluate on a single modality (like images) to emphasize that {\it multimodality helps unimodality}.

{\bf Cross-modal ensembles.} We now show that cross-modal learning produces classifiers that are ensembles of modality-specific classifiers, exposing a connection to related approaches for ensembling (such as WiSE-FT~\cite{wortsman2022robust}). We begin by appealing to the well-known {\it Representer Theorem}~\cite{scholkopf2001generalized}, which shows that optimally-trained classifiers can be represented as linear combinations of their training samples. In the case of a cross-modal linear probe, weights for class $y$ must be a weighted combination of all $i$ training features, across all modalities: 
\begin{align}
    w_y &= \sum_i \alpha_{iy} \phi_{m_i}(x_i) = \sum_{m \in M} w_y^m ,\quad \text{where} \nonumber\\ &w_y^m = \sum_{\{i: m_i = m\}} \alpha_{iy} \phi_m(x_i). \label{eq:ensemble}
\end{align} 
Linear classification via cross-modal adaptation solves for all weights $\alpha_{iy}$ {\em jointly}, so as to minimize the empirical risk (or training loss). In contrast, prior art optimizes for image-specific $\alpha_{iy}$'s {\em independently} of the text-specific $\alpha_{iy}$'s, linearly combining them with a single global $\alpha$ (as in WiSE-FT~\cite{wortsman2022robust}) or via text-based classifier initialization~\cite{zhang2021tip, gao2021clip}. Our analysis suggests that the joint optimization enabled by cross-modal learning may help other adaptation methods, as our experiments do in fact show.

{\bf Extensions.} Although we focus on uni-modal inference tasks (e.g. image classification), the above formulation allows the learned classifier to be applied to \textit{multimodal} test sets, such as classifying videos by training on image and audio, and then ensembling predictions across the two modalities with \autoref{eq:inference}. Or, one can extend image classification by providing additional data such as captions and/or attributes. We leave these scenarios as future work. Finally, just as one can optimize uni-modal losses \eqref{uni_modal_eq} by finetuning the encoder $\phi$, one can similarly finetune modality-specific encoders $\phi_m$ in the cross-modal setting~\eqref{cross_modal_eq}. We explore this finetuning method in the next section.

\section{Vision-Language Adaptation}

We now explore our cross-modal formulation for a particular multimodal setting. Many prior works~\cite{xing2019adaptive, mu2019shaping, zhou2022coop, zhang2021tip} explore the intersection of vision and language, and thus that is our initial focus. Interestingly, the influential ``zero-shot" and ``few-shot" evaluation protocols introduced by prior work~\cite{xian2018zero,radford2021learning} can be mapped to our cross-modal setting, with one crucial difference; the textual label of each class can be treated as an explicit training sample $(x_i,y_i,m_i)$. From this perspective, ``zero-shot" learning may be more naturally thought of as one-shot cross-modal learning that learns a few-shot model on {\em text} and then infers with it on {\em images}.

{\bf Few-shot evaluation protocol.} To ensure a fair comparison, we strictly follow the protocol of CoOp~\cite{zhou2022coop} by reporting test performance on 11 public image datasets (\autoref{tab:datasets}), with ResNet50~\cite{he2016deep} as the image encoder backbone. For maximal reproducibility, we use CoOp's dataset splits~\cite{zhou2022coop} and the three-fold few-shot train
sets sampled with the same random seeds. We adopt the given test split of each dataset %
as the test set. Some prior works~\cite{lu2022prompt, zhang2021tip} apparently use the large-scale test set to tune hyperparameters for few-shot learning; we instead exercise due diligence by tuning hyperparameters (such as the learning rate, weight decay, and early stopping) on the given few-shot validation set with $min(n, 4)$ examples, where $n$ is the number of training shots. We include PyTorch-style pseudocode (\autoref{algo:cross_modal_code}) and hyperparameter details (\autoref{sec:hyperparameters}).

\begin{algorithm}[t]
\scriptsize
\SetAlgoLined
    \PyComment{w: linear layer initialized with text features} \\
    \PyComment{T: temperature scaling (default is 100)} \\
    
    \PyCode{for \_ in iteration:}  \\
    \Indp
    \PyComment{Randomly sample images and texts} \\
    \PyCode{im, im\_labels = image\_loader.next()}  \\
    \PyCode{tx, tx\_labels = text\_loader.next()}  \\
    \PyCode{} \\
    \PyComment{Extract image and text features} \\
    \PyCode{im\_f = image\_encoder(im)}\\ 
    \PyCode{tx\_f = text\_encoder(tx)}\\ 
    \PyCode{} \\
    \PyComment{L2 normalize both features} \\
    \PyCode{im\_f = normalize(im\_f, dim=1)}\\ 
    \PyCode{tx\_f = normalize(tx\_f, dim=1)}\\ 
    \PyCode{} \\
    \PyComment{Compute softmax (cross entropy) loss} \\
    \PyCode{im\_loss = softmax\_loss(w(im\_f) / T, im\_labels)}\\ 
    \PyCode{tx\_loss = softmax\_loss(w(tx\_f) / T, tx\_labels)}\\ 
    \PyCode{loss = (im\_loss + tx\_loss) / 2}\\ 
    \PyCode{loss.backward()}\\ 
    \PyCode{}\\ 
    \PyComment{Update linear layer} \\
    \PyCode{update(w.params)} \\
    \PyComment{[optional] Update (partial or full) encoders} \\
    \PyCode{update(image\_encoder.params)} \\
    \PyCode{update(text\_encoder.params)} \\
    \Indm
\caption{An example of PyTorch-style pseudocode for cross-modal (vision-language) adaptation. Notably, the image and text samples do not need to be paired and one may sample different numbers of them per batch. For simplicity, we omit linear classifier initialization and early stopping with validation performance. One can also disable the corresponding {\tt grad} field of the encoders for partial finetuning, or pre-extract intermediate features to speed up training.}
\label{algo:cross_modal_code}
\end{algorithm}

\begin{table}[t]
\centering
\renewcommand{\arraystretch}{1.1}
\resizebox{\linewidth}{!}{
\begin{tabular}{c|ccccc|cc}
\toprule 
\multirow{2}{*}{Method} & \multicolumn{5}{c|}{Number of shots} & \multirow{2}{*}{Train speed}\\
\cmidrule(l){2-6} 
& 1 & 2 & 4 & 8 & 16 & \\
\midrule
{\bf Zero-Shot CLIP (58.8)} & - & - & - & - & - & - \\
Linear Probing & $36.7$ & $47.6$ & $57.2$ & $65.0$ & $71.1$ & \textcolor{green}{$<$1min} \\
WiSE-FT~\cite{wortsman2022robust} & $59.1$ & $61.8$ & $65.3$ & $68.4$ & $71.6$ & \textcolor{green}{$<$1min}\\
CoOp~\cite{zhou2022coop} & $59.6$ & $62.3$ & $66.8$ & $69.9$ & $73.4$ & \textcolor{red}{14hr} \\
ProGrad~\cite{zhu2022prompt} &  $62.6$ & $64.9$ & $68.5$ & $71.4$ & $74.0$ & \textcolor{red}{17hr} \\
Tip-Adapter~\cite{zhang2021tip} & $64.5$ & $66.7$ & $69.7$ & $72.5$ & $75.8$ & 5min \\
Tip-Adapter$^{\dagger}$
~\cite{zhang2021tip} & $63.3$ & $65.9$ & $69.0$ & $72.2$ & $75.1$ & 5min \\
 \hline
 
Cross-Modal Linear Probing & 64.1 & 67.0 & 70.3 & 73.0 & 76.0 & \textcolor{green}{$<$1min} \\
Cross-Modal Partial Finetuning & {\bf 64.7} & {\bf 67.2} & {\bf 70.5} & {\bf 73.6} & {\bf 77.1} & \textcolor{green}{$<$3min}\\
    \bottomrule
    \end{tabular}
}
\vspace{-2mm}
\caption{\small \textbf{Comparison to SOTA using the CoOp~\cite{zhou2022coop} protocol}, which reports top-1 accuracy across 11 test sets in~\autoref{tab:datasets}. We include per-dataset results and standard deviation in~\autoref{sec:all_results}. For a fair comparison, we reuse the same few-shot visual samples and hand-engineered text prompts used by Tip-Adapter~\cite{zhang2021tip}. The original Tip-Adapter searches over hyperparameters (e.g. early stopping) on the large-scale test set, which may not be realistic for few-shot scenarios. Instead, we rerun their \href{https://github.com/gaopengcuhk/Tip-Adapter/}{codebase} and early-stop on a few-shot validation set (as we do), denoted by $\dagger$. We reproduce WiSE-FT in our codebase since the original work does not provide few-shot results. In summary, by incorporating one-shot text samples into our training set, a simple cross-modal linear probe already outperforms {\em all} prior methods across {\em all} shots. Additionally, %
partial finetuning %
further improves performance, especially for 8 and 16 shots. Finally, our methods are faster to train than prior work, sometimes significantly (full report in~\autoref{tab:efficiency}).}
\label{tab:comparison_to_sota}
\end{table}

{\bf Cross-modal adaptation outperforms SOTA.} \autoref{tab:comparison_to_sota} shows the effectiveness of our proposal: we surpass all prior art with an embarrassingly simple linear classifier that requires significantly less training time than other carefully-crafted algorithms. In addition, partial finetuning of the last attentional pooling layer from $\phi_{image}$ sets the new SOTA. To ensure a fair comparison, we augment the class names into sentences using hand-engineered templates selected by Tip-Adapter~\cite{zhang2021tip} (\autoref{tab:datasets}) and follow their practice to initialize the linear layer with text features. Furthermore, we perform minimal image augmentation with a center crop plus a flipped view instead of random crops as in prior art~\cite{zhang2021tip, zhou2022coop}. As such, we can pre-extract features before training the classifier, leading to significantly less training time as shown in \autoref{tab:efficiency}. We also show that our method can benefit from both image and text augmentation in~\autoref{tab:augmentation}.  In the appendix, we provide more ablations on classifier initialization (\autoref{tab:init_results}), partial finetuning (\autoref{tab:partial_results}), and ViT-based backbone (\autoref{tab:complete_results}). Per-dataset results are also in appendix \autoref{tab:per_dataset}.

\begin{figure}
\vspace{-4mm}
    \centering
     \includegraphics[width=0.9\linewidth]{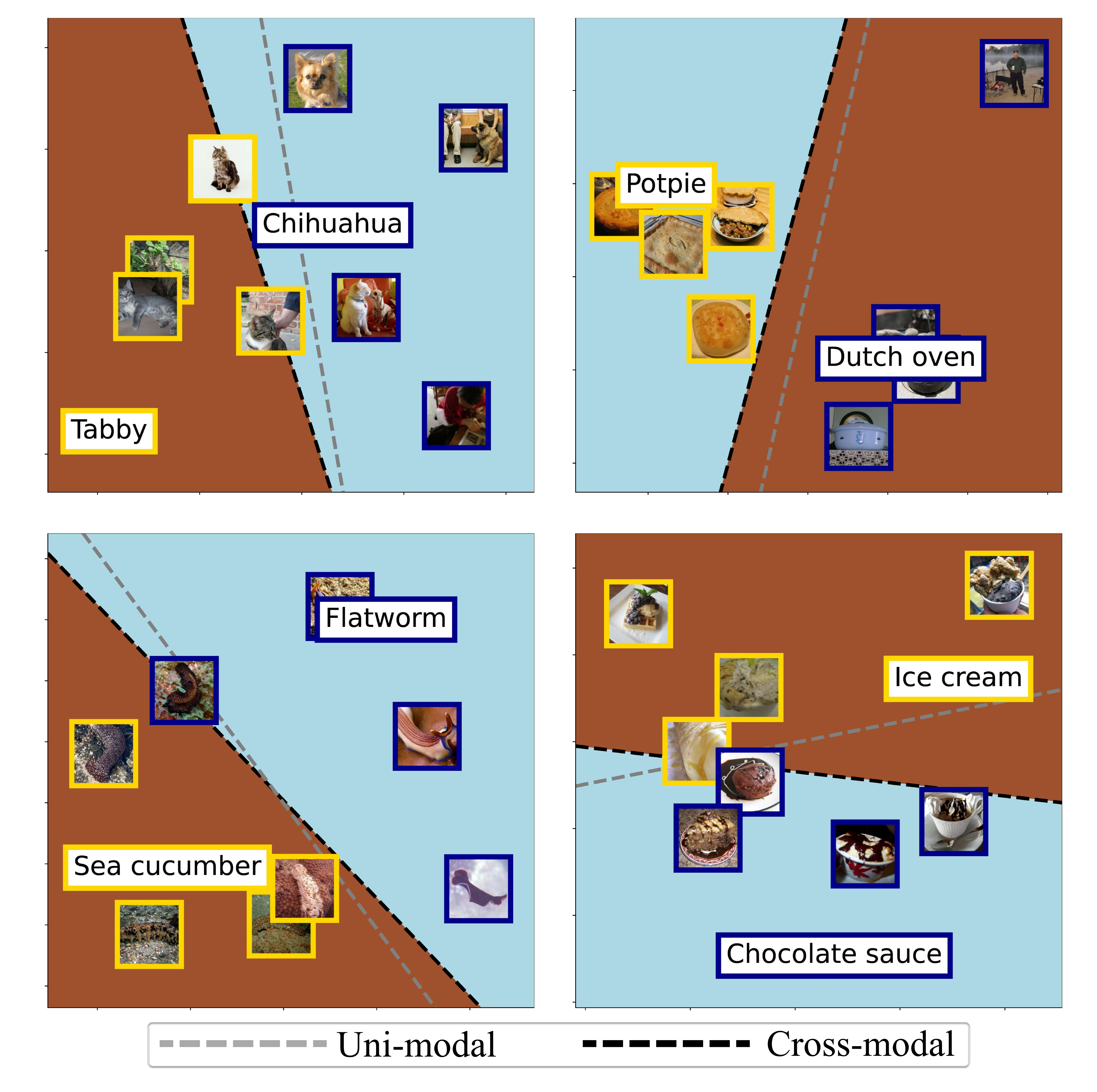}
     \vspace{-3mm}
    \caption{{\bf Additional PCA projection plots for random pairs of classes in ImageNet~\cite{deng2009imagenet}.} Adding one-shot text as training samples can oftentimes aggressively shift the decision boundary. }
    \vspace{-5mm}
    \label{fig:pcas}
\end{figure}

{\bf Why does cross-modal learning help?} As stated earlier, one reason that cross-modal learning helps is that it turns the original $n$-shot problem to an $(n+1)$-shot one.
However, \autoref{tab:comparison_to_sota} shows that  1-shot cross-modal linear probing outperforms the 2-shot results of most prior methods. This suggests that training samples from other modalities tend to contain complementary cues~\cite{wortsman2022robust, xing2019adaptive, mu2019shaping}. One can loosely observe this in \autoref{fig:pca_teaser} and \autoref{fig:pcas}, whereby visual and text examples lie in slightly different parts of the embedding space (indicating the potential to aggressively shape the final decision boundary). In fact, WiSE-FT~\cite{wortsman2022robust} is inspired by similar reasons to ensemble the uni-modal visual classifier with a ``zero-shot" (one-shot-text) classifier (in the linear probing case). However, \autoref{eq:ensemble} shows that cross-modal adaptation can also be seen as jointly learning an ensemble, while WiSE-FT~\cite{wortsman2022robust} learns the visual classifier independently of the text classifier. This suggests that other adaptation methods may benefit from cross-modal learning, as we show next.

\begin{table}[t]
\centering
\resizebox{\linewidth}{!}{
\begin{tabular}{c|ccccc}
\toprule 
\multirow{2}{*}{Method} & \multicolumn{5}{c}{Number of shots} \\
\cmidrule(l){2-6}
& 1 & 2 & 4 & 8 & 16 \\
\midrule
Linear Probing & $36.7$ & $47.6$ & $57.2$ & $65.0$ & $71.1$ \\
Cross-Modal Linear Probing & $64.1$ & $67.0$ & $70.3$ & $73.0$ & $76.0$ \\ 
$\Delta$ & \textcolor{green}{$27.4$} & \textcolor{green}{$19.4$} & \textcolor{green}{$13.1$} & \textcolor{green}{$8.0$} & \textcolor{green}{$4.9$}  \\
\hline
WiSE-FT~\cite{wortsman2022robust} & $59.1$ & $61.8$ & $65.3$ & $68.4$ & $71.6$ \\
Cross-Modal WiSE-FT & $63.8$ & $66.4$ & $69.0$ & $71.7$ & $74.1$ \\
$\Delta$ & \textcolor{green}{$4.7$} & \textcolor{green}{$4.6$} & \textcolor{green}{$3.7$} & \textcolor{green}{$3.3$} & \textcolor{green}{$2.5$}  \\
\hline
CoOp~\cite{zhou2022coop} & $59.6$ & $62.3$ & $66.8$ & $69.9$ & $73.4$  \\
Cross-Modal Prompting & $62.0$ & $64.9$ & $68.6$ & $71.4$ & $74.0$  \\
$\Delta$ & \textcolor{green}{$2.4$} & \textcolor{green}{$2.6$} & \textcolor{green}{$1.8$} & \textcolor{green}{$1.5$} & \textcolor{green}{$0.6$}  \\
\hline
Tip-Adapter$^{\dagger}$~\cite{zhang2021tip} & $63.3$ & $65.9$ & $69.0$ & $72.2$ & $75.1$\\
Cross-Modal Adapter & $64.4$ & $67.6$ & $70.8$ & $73.4$ & $75.9$  \\
$\Delta$ & \textcolor{green}{$1.1$} & \textcolor{green}{$1.7$} & \textcolor{green}{$1.8$} & \textcolor{green}{$1.2$} & \textcolor{green}{$0.8$}  \\
    \bottomrule
    \end{tabular}
}
\vspace{-2mm}
\caption{\small \textbf{Cross-modal adaptation improves existing methods.} We follow the same protocol as~\autoref{tab:comparison_to_sota}, reporting the delta accuracy between uni-modal and cross-modal variants of various state-of-the-art methods. The consistent boost suggests that cross-modal training is orthogonal to techniques for uni-modal adaptation, such as prompting~\cite{zhou2022coop}, adapter~\cite{houlsby2019parameter}, and robust finetuning~\cite{wortsman2022robust}.}
\label{tab:cross_modal_sota}
\end{table}

{\bf Cross-modal adaptation helps prior art (\autoref{tab:cross_modal_sota}).} This includes prompting (CoOp~\cite{zhou2022coop}), adapters (Tip-Adapter~\cite{zhang2021tip}), and robust-finetuning (WiSE-FT~\cite{wortsman2022robust}). We see a large improvement in the low-data regime (1 and 2 shots). Notably, we do not need to tune any methods, and simply reuse the reported hyperparameters. For prompting, we follow CoOp~\cite{zhou2022coop} to optimize 16 continuous tokens with the same training setting. For the Adapter model, we follow the same 2-layer MLP architecture of CLIP-Adapter~\cite{gao2021clip} with the given residual ratio of $0.2$; we outperform Tip-Adapter without relying on their training-free initialization of MLP. For WiSE-FT, we adopt the given ratio (0.5) to post-hoc ensemble the learned and the zero-shot classifiers. Overall, our experiments suggest that cross-modal adaptation is consistently effective, and should likely be a baseline moving forward given its ease-of-implementation (\autoref{algo:cross_modal_code}). For example, instead of separately benchmarking on ``zero-shot" (one-shot-text) and few-shot-vision, a cross-modal linear prob would suffice to evaluate representations of a multimodal model.%

\section{Vision-Audio Adaptation} %
\label{sec:generalize}

We now explore cross-modal adaption for other modalities such as audio. We pose the following question: can one learn a better dog  {\em visual} classifier by  {\em listening} to a dog barking? %
To examine this question, we curate the first audiovisual benchmark that supports few-shot classification of both image and audio.

{\bf Our ImageNet-ESC benchmark.}\footnote{Download instructions can be found in our \href{https://github.com/linzhiqiu/cross_modal_adaptation}{codebase}.} We construct our audiovisual benchmark by intersecting two of the most popular image and audio datasets: ImageNet~\cite{deng2009imagenet} with 1000 types of objects and ESC-50~\cite{piczak2015esc} with 50 types of environmental sounds (including animal, nature, human activity, domestic, and urban noises). We use the class names of the two datasets for class matching. For each class in ESC-50, we check whether there is a corresponding ImageNet class that may produce this type of sound. In this process, we observe that the audio-to-object matching can sometimes be one-to-many. For example, the {\tt clock-alarm} class in ESC-50 can be mapped to either {\tt digital clock} or {\tt analog clock} in ImageNet; the {\tt dog} (barking) class in ESC-50 can be matched to any of the 120 dog species. In such scenarios, we randomly match the classes, e.g. {\tt clock alarm} to {\tt digital clock} and {\tt dog} to {\tt otterhound}. Also, we find that some audio classes loosely match with some visual objects, such as {\tt drinking-sipping} to {\tt water bottle} and {\tt pouring-water} to {\tt water jug}. As such, we create two versions of the dataset: (1) {\bf ImageNet-ESC-27}, which represents the {\em maximal} intersection consisting of all loose matches, and (2) {\bf ImageNet-ESC-19}, a subset of the former version consisting of more accurate matches. The final matches are shown in appendix~\autoref{tab:imagenet_esc_dataset}.

{\bf Few-shot evaluation protocol. } We use five-fold few-shot splits sampled from ImageNet, with each split divided into half for training and validation. Test performance is recorded on the official ImageNet validation set of the corresponding classes. 
We adopt the predefined five folds of ESC-50, where each fold contains 8 samples per class. We construct 5 splits from ESC-50 by selecting one fold for training and validation, and record test performance on the other 4 folds. We report averaged performance over 25 runs (since we have 5 random splits for each modality). To keep consistent with our vision-language experiments, we adopt a uni-modal validation and test set and leave cross-modal testing for future work.

{\bf Audio encoding.} We use AudioCLIP~\cite{guzhov2021audioclip} with an ESResNeXT backbone~\cite{guzhov2021esresnextfbsp} as the audio encoder $\phi_{audio}$. Because AudioCLIP is trained on a large-scale video dataset (AudioSet~\cite{gemmeke2017audioset}) while  freezing the pre-trained CLIP text and image encoder, it produces audio embeddings in the same representation space. While AudioCLIP is pretrained on a sizable amount of data, we note that it does not come close to matching the scale of CLIP pretraining~\cite{guzhov2021audioclip,radford2021learning}. Thus, it does not perform favorably compared to the SOTA for downstream ``zero-shot" audio (i.e. one-shot text) classification tasks~\cite{guzhov2021audioclip}. However, scaling up audio pretraining is orthogonal to our investigation.

{\bf Audio improves image classification.} \autoref{tab:image_classification} shows that adding a random one-shot-audio improves upon naive image-only linear probing, especially in an extremely low-shot setting. This reaffirms ~\autoref{fig:ambiguity}'s hypothesis that cross-modality can reduce the ambiguity of the uni-modal few-shot setup; in other words, one can learn a better {\em image} classifier by {\em listen}ing to object sounds. One exception is the 4-shot performance on ImageNet-ESC-27, where adding audio does not help. We posit that (1) loosely-matched classes can result in noisier training data, and (2) the audio representations are not as robust due to smaller-scale pretraining. This suggests that cross-modal adaptation is less effective when representations are not aligned well or insufficiently trained. Nevertheless, under most scenarios, cross-modal adaptation helps. \autoref{tab:image_complete} shows that adding the language modality (i.e. label names) can significantly boost the performance, which is expected because our benchmark is curated with textual information. For all experiments, we follow an identical procedure to vision-language experiments in \autoref{sec:vision-language} and provide details in appendix~\autoref{sec:hyperparameters}.

{\bf Vision improves audio classification.} We additionally evaluate the \textit{reverse} task -- whether adding a random one-shot \textit{image} sample for downstream audio classification can improve upon audio-only training. 
\autoref{tab:audio_classification} shows the results, where we see the same favorable trend. This success concludes that our approachis modality-agnostic.

\begin{table}[t]
\centering
\renewcommand{\arraystretch}{1.2}
\resizebox{\linewidth}{!}{
\begin{tabular}{cc|ccc}
\toprule 
\multirow{2}{*}{Dataset} & \multirow{2}{*}{Method} & \multicolumn{3}{c}{Image Classification}  \\
\cmidrule(l){3-5}
& & 1-shot & 2-shot & 4-shot \\
\midrule
\multirow{2}{*}{ImageNet-ESC-19} & Image-Only Linear & 68.0 & 75.7 & 83.1 \\
& Image-Audio Linear & {\bf 69.3} & {\bf 76.7} &  {\bf 83.2} \\
\hline
\multirow{2}{*}{ImageNet-ESC-27} & Image-Only Linear & $60.1$ & $71.8$ & {\bf 79.0}  \\
& Image-Audio Linear &  {\bf 60.9}	& {\bf 73.3} & 78.9 \\
    \bottomrule
    \end{tabular}
}
\vspace{-2mm}
\caption{{\bf Image classification results on ImageNet-ESC benchmark.} Adding one audio shot can improve image classification under most few-shot scenarios, even when the audio and vision modalities are only loosely aligned. \vspace{-1mm}}
\label{tab:image_classification}
\end{table}

\begin{table}[t]
\centering
\renewcommand{\arraystretch}{1.2}
\resizebox{\linewidth}{!}{
\begin{tabular}{cc|ccc}
\toprule 
\multirow{2}{*}{Dataset} & \multirow{2}{*}{Method} & \multicolumn{3}{c}{Audio Classification}  \\
\cmidrule(l){3-5}
& & 1-shot & 2-shot & 4-shot \\
\midrule
\multirow{2}{*}{ImageNet-ESC-19} & Audio-Only Linear & 31.2 & 41.1 & 48.5 \\ 
& Audio-Image Linear & {\bf 35.7} & {\bf 45.9} &  {\bf 51.6} \\ 
\hline
\multirow{2}{*}{ImageNet-ESC-27} & Audio-Only Linear & $28.2$	& $39.0$ & $47.1$\\
& Audio-Image Linear &  {\bf 35.0}	& {\bf 43.5} & {\bf 48.5} \\
    \bottomrule
    \end{tabular}
}
\vspace{-2mm}
\caption{{\bf Audio classification results on ImageNet-ESC benchmark.} Similar to~\autoref{tab:image_classification}, adding one image shot improves few-shot audio classification. \vspace{-1mm}}
\label{tab:audio_classification}
\end{table}

\section{Ablation Studies}

We present a few selected ablation studies in this section. For comprehensive results, please refer to~\autoref{sec:all_results}.

\begin{table}[t]
\centering
\renewcommand{\arraystretch}{1.2}
\resizebox{\linewidth}{!}{
\begin{tabular}{cccccc}
\toprule 
Dataset & Classes & Train & Val & Test & Hand-crafted Prompt~\cite{zhang2021tip}\\
\midrule
Caltech101~\cite{fei2004learning} & 100& 4,128& 1,649& 2,465 & {\tt a photo of a \{cls\}.} \\
OxfordPets~\cite{parkhi2012cats} & 37 & 2,944 & 736 & 3,669 & {\tt a photo of a \{cls\}, a type of pet.} \\
StanfordCars~\cite{krause20133d} & 196 & 6,509 & 1,635 & 8,041 & {\tt a photo of a \{cls\}.} \\
Flowers102~\cite{nilsback2008automated} & 102 & 4,093 & 1,633 & 2,463 & {\tt a photo of a \{cls\}, a type of flower.} \\
Food101~\cite{bossard2014food} & 101 & 50,500 & 20,200 & 30,300 & {\tt a photo of \{cls\}, a type of food.} \\
FGVCAircraft~\cite{maji2013fine} & 100 & 3,334 & 3,333 & 3,333 & {\tt a photo of a \{cls\}, a type of aircraft.} \\
SUN397~\cite{xiao2010sun} & 397 & 15,880 & 3,970 & 19,850 & {\tt a photo of a \{cls\}.} \\
DTD~\cite{cimpoi2014describing} & 47 & 2,820 & 1,128 & 1,692 & {\tt \{cls\} texture.} \\
EuroSAT~\cite{helber2019eurosat} & 10 & 13,500 & 5,400 & 8,100 & {\tt a centered satellite photo of \{cls\}.} \\
UCF101~\cite{soomro2012ucf101} & 101 & 7,639 & 1,898 & 3,783 & {\tt a photo of a person doing \{cls\}.} \\
\hline
ImageNet~\cite{deng2009imagenet} & 1000 & 1.28M & N/A & 50,000 & \shortstack{{\tt itap of a \{cls\}.} \\
{\tt a bad photo of the \{cls\}.} \\
{\tt a origami \{cls\}.} \\
{\tt a photo of the large \{cls\}.} \\
{\tt a \{cls\} in a video game.} \\
{\tt art of the \{cls\}.} \\
{\tt a photo of the small \{cls\}.}}\\
    \bottomrule
    \end{tabular}
}
\vspace{-2mm}
\caption{\small \textbf{Detailed statistics of the 11 datasets. } We adopt the hand-engineered templates selected by Tip-Adapter~\cite{zhang2021tip} unless otherwise stated. Note that this set of templates is identical to the ones selected by CLIP~\cite{radford2021learning} and CoOp~\cite{zhou2022coop}, except for ImageNet.\vspace{-4mm}}
\label{tab:datasets}
\end{table}

{\bf Data augmentation of text samples. } Like most prior works~\cite{radford2021learning, zhou2022coop}, we also find that data augmentation can improve downstream performance during vision-language adaptation (cf.~\autoref{tab:comparison_to_sota}). Notably, since the class names are included as training samples, one can explore augmentation techniques for text (just as random cropping for images). Besides the fixed template {\tt a photo of a \{cls\}} and hand-crafted templates (\autoref{tab:datasets}), we also try a {\bf template mining} strategy that does not rely on the selected dataset-specific templates. To automatically mine for the templates, we search among a pool of 180 templates for 21 templates with the best zero-shot performance on the few-shot validation set of each dataset.  We discuss how we collect the 180 templates in appendix~\autoref{sec:hyperparameters}. For image augmentation, we perform standard flipping and random cropping. We show a subset of results in \autoref{tab:augmentation}, and find that all text augmentation techniques provide a sizable boost in performance. We also report comprehensive ablations in appendix~\autoref{tab:augmentation_all} and compare it to the SOTA prompting method ProDA~\cite{lu2022prompt}. The salient conclusions include (1) the performance gain from image augmentation is saturated after more than two views, and (2) template mining can be as competitive as a large number of 36 carefully-tuned prompts. In fact, prompting~\cite{zhou2022coop, lu2022prompt, liu2021pre} can be viewed as another {\em text augmentation} technique under cross-modal adaptation, and we leave this exploration to future work.

\begin{table}
\centering
\renewcommand{\arraystretch}{1.2}
\resizebox{\linewidth}{!}{
\begin{tabular}{ccc|ccccc}
\toprule 
\multirow{2}{*}{Finetuning} & \multirow{2}{*}{ImageAugment} & \multirow{2}{*}{TextAugment} & \multicolumn{5}{c}{Number of shots} \\
\cmidrule(l){4-8}
& & & 1 & 2 & 4 & 8 & 16 \\
\midrule
\multirow{5}{*}{Linear} & \multirow{4}{*}{CenterCrop} & Classname & 61.8 & 65.3 & 69.0 & 72.0 & 74.9 \\
&  & {\tt a photo of a \{cls\}.} & 63.2 & 66.2 & 69.7 & 72.5 & 75.3 \\
& & Template Mining & 63.5 & 67.2 & 70.3 & 73.1 & 75.7 \\
& & Hand Engineered~\cite{zhang2021tip} & 63.7 & 66.7 & 70.3 & 72.9 & 75.5 \\
\cmidrule(l){2-8}
& +Flipped View & Hand Engineered~\cite{zhang2021tip} & {\bf 64.1} & {\bf 67.0} & {\bf 70.3} & {\bf 73.0} & {\bf 76.0} \\
\hline
\multirow{5}{*}{Partial} & \multirow{4}{*}{CenterCrop} & Classname & 62.5 & 65.7 & 69.3 & 72.9 & 76.2 \\
&  & {\tt a photo of a \{cls\}.} & 63.8 & 66.8 & 69.8 & 73.4 & 76.7 \\
& & Template Mining & 64.3 & 67.1 & 70.3 & 73.5 & 76.5 \\
& & Hand Engineered~\cite{zhang2021tip} & 64.6 & 67.2 & 70.2 & 73.7 & 76.9 \\
\cmidrule(l){2-8}
& +Flipped View & Hand Engineered~\cite{zhang2021tip} & {\bf 64.7} & {\bf 67.7} & {\bf 70.6} & {\bf 73.8} & {\bf 77.2} \\
    \bottomrule
    \end{tabular}
}
\vspace{-2mm}
\caption{\small \textbf{Augmentation for cross-modal adaptation.} We evaluate the impact of selected augmentation techniques following the same CoOp protocol as in \autoref{tab:comparison_to_sota}. }
\vspace{-5mm}
\label{tab:augmentation}
\end{table}

{\bf Test-time distribution shifts.} We examine how robust our approach is against test-time distribution shifts in \autoref{tab:robustness}. Specifically, we follow the CoOp~\cite{zhou2022coop} protocol to report the test performance of a classifier trained on the source dataset (16-shot ImageNet) to 4 distribution-shifted target test sets, including ImageNet-V2~\cite{recht2019imagenet}, ImageNet-Sketch~\cite{wang2019learning}, ImageNet-A~\cite{hendrycks2021natural}, and ImageNet-R~\cite{hendrycks2021many}. As shown in \autoref{tab:robustness}, cross-modal adaptation can significantly boost the robustness of image-only linear probing and is competitive against baselines designed to address robustness such as CoCoOp~\cite{zhou2022cocoop} and WiSE-FT~\cite{wortsman2022robust}. Cross-Modal adaptation also improves upon WiSE-FT~\cite{wortsman2022robust} and sets the new SOTA. We can conclude that language modality plays an important role in robustness, similar to how humans rely on textual cues for recognition~\cite{hendrycks2021natural}.

\begin{table}[t]
\centering
\renewcommand{\arraystretch}{1.1}
\resizebox{\linewidth}{!}{
\begin{tabular}{c|ccccc}
\toprule 
\multirow{2}{*}{Method} & Source & \multicolumn{4}{c}{Target} \\
\cmidrule(l){2-2} \cmidrule(l){3-6}
& ImageNet & -V2 & -Sketch & -A & -R \\
\midrule
\multicolumn{1}{l}{{\bf ResNet50}} \\
Zero-Shot CLIP & 58.2 & 51.3 & 33.3 & 21.7 & 56.0\\
Linear Probing & 55.9 & 46.0 & 19.1 & 12.7 & 34.9\\
CoOp (M=4) & 63.0 & 55.1 & 32.7 & 22.1 & 55.0 \\
CoOp (M=16) & 63.3 & \underline{55.4} & \underline{34.7} & \textbf{23.1} & 56.6 \\
WiSE-FT ($\alpha$=0.5) & 62.9 & 54.2 & 33.3 & 20.3 & \underline{57.4}\\
Cross-Modal WiSE-FT ($\alpha$=0.5) & \textbf{65.2} & \textbf{56.6} & \textbf{35.6} & \underline{22.6} & \textbf{59.5} \\
Cross-Modal Linear Probing &  \underline{64.5} & 55.3 & 33.1 & 20.0 & 56.4 \\
\hline
\multicolumn{1}{l}{{\bf ViT-B/16}} \\
Zero-Shot CLIP &  66.7 & 60.8 & 46.2 & 47.8 & 74.0 \\
Linear Probing & 65.9 & 56.3 & 34.8 & 35.7 & 58.4 \\
CoOp (M=4) & 71.9 & 64.2 & 46.7 & 48.4 & 74.3\\
CoOp (M=16) & 71.7 & 64.6 & 47.9 & 49.9 & 75.1\\
CoCoOp & 71.0 & 64.1 & 48.8 & \textbf{50.6} & 76.2 \\
WiSE-FT ($\alpha$=0.5) & \underline{73.0} & \underline{65.2} & \underline{49.1} & 49.8 & \underline{77.6} \\
Cross-Modal WiSE-FT ($\alpha$=0.5) & 72.9 & \textbf{65.4} & \textbf{49.2} & \underline{50.5} &  \textbf{77.8} \\
Cross-Modal Linear Probing & \textbf{73.2} & 64.8 & 47.9 & 48.3 & 76.4 \\
    \bottomrule
    \end{tabular}
}
\vspace{-2mm}
\caption{{\bf Robustness under test-time distribution shifts.} We follow CoOp~\cite{zhou2022coop}'s protocol for evaluating the test-time performance on variants of ImageNet. We report results with two image encoders (ResNet50 and ViT-B/16), and mark the \textbf{best} and \underline{second best} results. Salient conclusions: (a) Cross-modal linear probing is much more robust than its uni-modal counterpart while being competitive to previous SOTA methods such as WiseFT and CoOp, and (b) it can be further augmented with post-hoc modification through WiseFT to achieve new the SOTA.}
\label{tab:robustness}
\end{table}

{\bf Efficiency.} As shown in \autoref{tab:efficiency}, our approaches are much more lightweight because we do not rely on deep finetuning~\cite{zhou2022coop, zhou2022cocoop} or heavy image augmentations. This allows us to speed up training by pre-extracting features, resulting in rather fast training speeds.

\begin{table}[t]
\vspace{-3mm}
\centering
\renewcommand{\arraystretch}{1.1}
\resizebox{\linewidth}{!}{
\begin{tabular}{cccccccc}
\toprule 
Method & Iteration & Time & Accuracy & Gain \\
\midrule
Zero-shot CLIP~\cite{radford2021learning} & 0 & 0 & 60.33 & 0 \\
Image-Only Linear &  12k & {\bf 15sec} & 56.44 & -3.89 \\
CoOp~\cite{zhou2022coop} &  100k & 14h 40min & 62.95  & +2.62 \\
ProGrad~\cite{zhou2022coop} &  100k & 17hr & 63.45  & +3.12 \\
Tip-Adapter~\cite{zhang2021tip} &  10k & 5min & 65.18 & +5.18 \\
\hline
Cross-Modal Linear &  12k & {\bf 15sec} & 64.51 & +4.14 \\
Cross-Modal Partial &  12k & 2.5min & {\bf 65.95} & {\bf +5.57} \\
    \bottomrule
    \end{tabular}
}
\vspace{-2mm}
\caption{\small \textbf{Efficiency and accuracy for different methods on ImageNet-16-shot.} All experiments are tested with batch size 32 on a single NVIDIA GeForce RTX 3090 GPU. Our approaches take less time and achieve SOTA performance. }
\vspace{-4mm}
\label{tab:efficiency}
\end{table}

\section{Discussion and Limitations}
We show that cross-modal training is a lightweight and effective approach for adapting pre-trained  multimodal models for downstream uni-modal tasks. One reason for its effectiveness is that it naturally addresses the underspecification of few-shot learning. In the context of vision-language adaptation, one can achieve SOTA results by using existing text labels as free training samples. In the context of vision-audio adapation, one can learn better visual object classifiers by listening to object sounds (and better audio classifiers by looking at objects!).  %
One attractive aspect of cross-modal learning is that the learned models naturally apply to multimodal test data, such as the classification of videos that contain both visual and audio signals. However, cross-modal learning is less effective when model representations are not well-aligned or insufficiently trained. Nevertheless, due to its simplicity and effectiveness, we hope cross-modal learning becomes a tool for future research on multi-modal adaptation.

{\small
\bibliographystyle{ieee_fullname}
\bibliography{egbib}

\begin{thebibliography}{100}\itemsep=-1pt

\bibitem{afham2021rich}
Mohamed Afham, Salman Khan, Muhammad~Haris Khan, Muzammal Naseer, and
  Fahad~Shahbaz Khan.
\newblock Rich semantics improve few-shot learning.
\newblock {\em arXiv preprint arXiv:2104.12709}, 2021.

\bibitem{alayrac2022flamingo}
Jean-Baptiste Alayrac, Jeff Donahue, Pauline Luc, Antoine Miech, Iain Barr,
  Yana Hasson, Karel Lenc, Arthur Mensch, Katie Millican, Malcolm Reynolds,
  et~al.
\newblock Flamingo: a visual language model for few-shot learning.
\newblock {\em arXiv preprint arXiv:2204.14198}, 2022.

\bibitem{alwassel2020self}
Humam Alwassel, Dhruv Mahajan, Bruno Korbar, Lorenzo Torresani, Bernard Ghanem,
  and Du Tran.
\newblock Self-supervised learning by cross-modal audio-video clustering.
\newblock {\em Advances in Neural Information Processing Systems},
  33:9758--9770, 2020.

\bibitem{bateni2020improved}
Peyman Bateni, Raghav Goyal, Vaden Masrani, Frank Wood, and Leonid Sigal.
\newblock Improved few-shot visual classification.
\newblock In {\em Proceedings of the IEEE/CVF Conference on Computer Vision and
  Pattern Recognition}, pages 14493--14502, 2020.

\bibitem{bommasani2021opportunities}
Rishi Bommasani, Drew~A Hudson, Ehsan Adeli, Russ Altman, Simran Arora, Sydney
  von Arx, Michael~S Bernstein, Jeannette Bohg, Antoine Bosselut, Emma
  Brunskill, et~al.
\newblock On the opportunities and risks of foundation models.
\newblock {\em arXiv preprint arXiv:2108.07258}, 2021.

\bibitem{bossard2014food}
Lukas Bossard, Matthieu Guillaumin, and Luc~Van Gool.
\newblock Food-101--mining discriminative components with random forests.
\newblock In {\em European conference on computer vision}, pages 446--461.
  Springer, 2014.

\bibitem{bossard14food}
Lukas Bossard, Matthieu Guillaumin, and Luc Van~Gool.
\newblock Food-101 -- mining discriminative components with random forests.
\newblock In {\em European Conference on Computer Vision}, 2014.

\bibitem{brown2020gpt3}
Tom Brown, Benjamin Mann, Nick Ryder, Melanie Subbiah, Jared~D Kaplan, Prafulla
  Dhariwal, Arvind Neelakantan, Pranav Shyam, Girish Sastry, Amanda Askell,
  Sandhini Agarwal, Ariel Herbert-Voss, Gretchen Krueger, Tom Henighan, Rewon
  Child, Aditya Ramesh, Daniel Ziegler, Jeffrey Wu, Clemens Winter, Chris
  Hesse, Mark Chen, Eric Sigler, Mateusz Litwin, Scott Gray, Benjamin Chess,
  Jack Clark, Christopher Berner, Sam McCandlish, Alec Radford, Ilya Sutskever,
  and Dario Amodei.
\newblock Language models are few-shot learners.
\newblock In H. Larochelle, M. Ranzato, R. Hadsell, M.F. Balcan, and H. Lin,
  editors, {\em Advances in Neural Information Processing Systems}, volume~33,
  pages 1877--1901. Curran Associates, Inc., 2020.

\bibitem{Calvert1997lip}
Gemma Calvert, Edward Bullmore, M.J. Brammer, Ruth Campbell, Steven Williams,
  Philip Mcguire, Peter Woodruff, S.D. Iversen, and Anthony David.
\newblock Activation of auditory cortex during silent lipreading. science,
  276(5312), 593-596.
\newblock {\em Science (New York, N.Y.)}, 276:593--6, 05 1997.

\bibitem{cangea2019xflow}
C{\u{a}}t{\u{a}}lina Cangea, Petar Veli{\v{c}}kovi{\'c}, and Pietro Lio.
\newblock Xflow: Cross-modal deep neural networks for audiovisual
  classification.
\newblock {\em IEEE Transactions on Neural Networks and Learning Systems},
  31(9):3711--3720, 2019.

\bibitem{caron2021emerging}
Mathilde Caron, Hugo Touvron, Ishan Misra, Herv{\'e} J{\'e}gou, Julien Mairal,
  Piotr Bojanowski, and Armand Joulin.
\newblock Emerging properties in self-supervised vision transformers.
\newblock In {\em Proceedings of the IEEE/CVF International Conference on
  Computer Vision}, pages 9650--9660, 2021.

\bibitem{chen2020simple}
Ting Chen, Simon Kornblith, Mohammad Norouzi, and Geoffrey Hinton.
\newblock A simple framework for contrastive learning of visual
  representations.
\newblock In {\em International conference on machine learning}, pages
  1597--1607. PMLR, 2020.

\bibitem{cimpoi14dtd}
M. Cimpoi, S. Maji, I. Kokkinos, S. Mohamed, , and A. Vedaldi.
\newblock Describing textures in the wild.
\newblock In {\em Proceedings of the {IEEE} Conf. on Computer Vision and
  Pattern Recognition ({CVPR})}, 2014.

\bibitem{cimpoi2014describing}
Mircea Cimpoi, Subhransu Maji, Iasonas Kokkinos, Sammy Mohamed, and Andrea
  Vedaldi.
\newblock Describing textures in the wild.
\newblock In {\em Proceedings of the IEEE conference on computer vision and
  pattern recognition}, pages 3606--3613, 2014.

\bibitem{deng2009imagenet}
Jia Deng, Wei Dong, Richard Socher, Li-Jia Li, Kai Li, and Li Fei-Fei.
\newblock Imagenet: A large-scale hierarchical image database.
\newblock In {\em 2009 IEEE conference on computer vision and pattern
  recognition}, pages 248--255. Ieee, 2009.

\bibitem{deng2022rlprompt}
Mingkai Deng, Jianyu Wang, Cheng-Ping Hsieh, Yihan Wang, Han Guo, Tianmin Shu,
  Meng Song, Eric~P Xing, and Zhiting Hu.
\newblock Rlprompt: Optimizing discrete text prompts with reinforcement
  learning.
\newblock {\em arXiv preprint arXiv:2205.12548}, 2022.

\bibitem{devlin2018bert}
Jacob Devlin, Ming-Wei Chang, Kenton Lee, and Kristina Toutanova.
\newblock Bert: Pre-training of deep bidirectional transformers for language
  understanding.
\newblock {\em arXiv preprint arXiv:1810.04805}, 2018.

\bibitem{dhillon2019baseline}
Guneet~S Dhillon, Pratik Chaudhari, Avinash Ravichandran, and Stefano Soatto.
\newblock A baseline for few-shot image classification.
\newblock {\em arXiv preprint arXiv:1909.02729}, 2019.

\bibitem{fei2004learning}
Li Fei-Fei, Rob Fergus, and Pietro Perona.
\newblock Learning generative visual models from few training examples: An
  incremental bayesian approach tested on 101 object categories.
\newblock In {\em 2004 conference on computer vision and pattern recognition
  workshop}, pages 178--178. IEEE, 2004.

\bibitem{finn2017model}
Chelsea Finn, Pieter Abbeel, and Sergey Levine.
\newblock Model-agnostic meta-learning for fast adaptation of deep networks.
\newblock In {\em International conference on machine learning}, pages
  1126--1135. PMLR, 2017.

\bibitem{gao2021clip}
Peng Gao, Shijie Geng, Renrui Zhang, Teli Ma, Rongyao Fang, Yongfeng Zhang,
  Hongsheng Li, and Yu Qiao.
\newblock Clip-adapter: Better vision-language models with feature adapters.
\newblock {\em arXiv preprint arXiv:2110.04544}, 2021.

\bibitem{gao2020making}
Tianyu Gao, Adam Fisch, and Danqi Chen.
\newblock Making pre-trained language models better few-shot learners.
\newblock {\em arXiv preprint arXiv:2012.15723}, 2020.

\bibitem{gemmeke2017audioset}
Jort~F. Gemmeke, Daniel P.~W. Ellis, Dylan Freedman, Aren Jansen, Wade
  Lawrence, R.~Channing Moore, Manoj Plakal, and Marvin Ritter.
\newblock Audio set: An ontology and human-labeled dataset for audio events.
\newblock In {\em 2017 IEEE International Conference on Acoustics, Speech and
  Signal Processing (ICASSP)}, pages 776--780, 2017.

\bibitem{gibson1969principles}
Eleanor~J Gibson.
\newblock Principles of perceptual learning and development.
\newblock 1969.

\bibitem{girdhar2017attentional}
Rohit Girdhar and Deva Ramanan.
\newblock Attentional pooling for action recognition.
\newblock {\em Advances in neural information processing systems}, 30, 2017.

\bibitem{goyal2021self}
Priya Goyal, Mathilde Caron, Benjamin Lefaudeux, Min Xu, Pengchao Wang, Vivek
  Pai, Mannat Singh, Vitaliy Liptchinsky, Ishan Misra, Armand Joulin, et~al.
\newblock Self-supervised pretraining of visual features in the wild.
\newblock {\em arXiv preprint arXiv:2103.01988}, 2021.

\bibitem{guzhov2021audioclip}
Andrey Guzhov, Federico Raue, Jörn Hees, and Andreas Dengel.
\newblock Audioclip: Extending clip to image, text and audio, 2021.

\bibitem{guzhov2021esresnextfbsp}
Andrey Guzhov, Federico Raue, Jörn Hees, and Andreas Dengel.
\newblock Esresne(x)t-fbsp: Learning robust time-frequency transformation of
  audio, 2021.

\bibitem{hariharan2017low}
Bharath Hariharan and Ross Girshick.
\newblock Low-shot visual recognition by shrinking and hallucinating features.
\newblock In {\em Proceedings of the IEEE international conference on computer
  vision}, pages 3018--3027, 2017.

\bibitem{haviv2021bertese}
Adi Haviv, Jonathan Berant, and Amir Globerson.
\newblock Bertese: Learning to speak to bert.
\newblock {\em arXiv preprint arXiv:2103.05327}, 2021.

\bibitem{he2022masked}
Kaiming He, Xinlei Chen, Saining Xie, Yanghao Li, Piotr Doll{\'a}r, and Ross
  Girshick.
\newblock Masked autoencoders are scalable vision learners.
\newblock In {\em Proceedings of the IEEE/CVF Conference on Computer Vision and
  Pattern Recognition}, pages 16000--16009, 2022.

\bibitem{he2020momentum}
Kaiming He, Haoqi Fan, Yuxin Wu, Saining Xie, and Ross Girshick.
\newblock Momentum contrast for unsupervised visual representation learning.
\newblock In {\em Proceedings of the IEEE/CVF conference on computer vision and
  pattern recognition}, pages 9729--9738, 2020.

\bibitem{he2016deep}
Kaiming He, Xiangyu Zhang, Shaoqing Ren, and Jian Sun.
\newblock Deep residual learning for image recognition.
\newblock In {\em Proceedings of the IEEE conference on computer vision and
  pattern recognition}, pages 770--778, 2016.

\bibitem{helber2017eurosat}
Patrick Helber, Benjamin Bischke, Andreas Dengel, and Damian Borth.
\newblock Eurosat: A novel dataset and deep learning benchmark for land use and
  land cover classification, 2017.

\bibitem{helber2019eurosat}
Patrick Helber, Benjamin Bischke, Andreas Dengel, and Damian Borth.
\newblock Eurosat: A novel dataset and deep learning benchmark for land use and
  land cover classification.
\newblock {\em IEEE Journal of Selected Topics in Applied Earth Observations
  and Remote Sensing}, 12(7):2217--2226, 2019.

\bibitem{hendrycks2021many}
Dan Hendrycks, Steven Basart, Norman Mu, Saurav Kadavath, Frank Wang, Evan
  Dorundo, Rahul Desai, Tyler Zhu, Samyak Parajuli, Mike Guo, Dawn Song, Jacob
  Steinhardt, and Justin Gilmer.
\newblock The many faces of robustness: A critical analysis of
  out-of-distribution generalization.
\newblock {\em ICCV}, 2021.

\bibitem{hendrycks2021natural}
Dan Hendrycks, Kevin Zhao, Steven Basart, Jacob Steinhardt, and Dawn Song.
\newblock Natural adversarial examples.
\newblock In {\em Proceedings of the IEEE/CVF Conference on Computer Vision and
  Pattern Recognition}, pages 15262--15271, 2021.

\bibitem{hong2020x}
Danfeng Hong, Naoto Yokoya, Gui-Song Xia, Jocelyn Chanussot, and Xiao~Xiang
  Zhu.
\newblock X-modalnet: A semi-supervised deep cross-modal network for
  classification of remote sensing data.
\newblock {\em ISPRS Journal of Photogrammetry and Remote Sensing}, 167:12--23,
  2020.

\bibitem{houlsby2019parameter}
Neil Houlsby, Andrei Giurgiu, Stanislaw Jastrzebski, Bruna Morrone, Quentin
  De~Laroussilhe, Andrea Gesmundo, Mona Attariyan, and Sylvain Gelly.
\newblock Parameter-efficient transfer learning for nlp.
\newblock In {\em International Conference on Machine Learning}, pages
  2790--2799. PMLR, 2019.

\bibitem{huang2022unsupervised}
Tony Huang, Jack Chu, and Fangyun Wei.
\newblock Unsupervised prompt learning for vision-language models.
\newblock {\em arXiv preprint arXiv:2204.03649}, 2022.

\bibitem{ilharco2022patching}
Gabriel Ilharco, Mitchell Wortsman, Samir~Yitzhak Gadre, Shuran Song, Hannaneh
  Hajishirzi, Simon Kornblith, Ali Farhadi, and Ludwig Schmidt.
\newblock Patching open-vocabulary models by interpolating weights.
\newblock {\em arXiv preprint arXiv:2208.05592}, 2022.

\bibitem{jackendoff1987beyond}
Ray Jackendoff.
\newblock On beyond zebra: The relation of linguistic and visual information.
\newblock {\em Cognition}, 26(2):89--114, 1987.

\bibitem{jia2021scaling}
Chao Jia, Yinfei Yang, Ye Xia, Yi-Ting Chen, Zarana Parekh, Hieu Pham, Quoc Le,
  Yun-Hsuan Sung, Zhen Li, and Tom Duerig.
\newblock Scaling up visual and vision-language representation learning with
  noisy text supervision.
\newblock In {\em International Conference on Machine Learning}, pages
  4904--4916. PMLR, 2021.

\bibitem{jia2022visual}
Menglin Jia, Luming Tang, Bor-Chun Chen, Claire Cardie, Serge Belongie, Bharath
  Hariharan, and Ser-Nam Lim.
\newblock Visual prompt tuning.
\newblock {\em arXiv preprint arXiv:2203.12119}, 2022.

\bibitem{jiang2020can}
Zhengbao Jiang, Frank~F Xu, Jun Araki, and Graham Neubig.
\newblock How can we know what language models know?
\newblock {\em Transactions of the Association for Computational Linguistics},
  8:423--438, 2020.

\bibitem{joachims1999transductive}
Thorsten Joachims et~al.
\newblock Transductive inference for text classification using support vector
  machines.
\newblock In {\em Icml}, volume~99, pages 200--209, 1999.

\bibitem{joulin2016learning}
Armand Joulin, Laurens van~der Maaten, Allan Jabri, and Nicolas Vasilache.
\newblock Learning visual features from large weakly supervised data.
\newblock In {\em European Conference on Computer Vision}, pages 67--84.
  Springer, 2016.

\bibitem{kirkpatrick2017overcoming}
James Kirkpatrick, Razvan Pascanu, Neil Rabinowitz, Joel Veness, Guillaume
  Desjardins, Andrei~A Rusu, Kieran Milan, John Quan, Tiago Ramalho, Agnieszka
  Grabska-Barwinska, et~al.
\newblock Overcoming catastrophic forgetting in neural networks.
\newblock {\em Proceedings of the national academy of sciences},
  114(13):3521--3526, 2017.

\bibitem{kosslyn2010brainimagery}
Stephen~M. Kosslyn, Giorgio Ganis, and William~L. Thompson.
\newblock {3Multimodal images in the brain}.
\newblock In {\em {The neurophysiological foundations of mental and motor
  imagery}}. Oxford University Press, 01 2010.

\bibitem{krause20133d}
Jonathan Krause, Michael Stark, Jia Deng, and Li Fei-Fei.
\newblock 3d object representations for fine-grained categorization.
\newblock In {\em Proceedings of the IEEE international conference on computer
  vision workshops}, pages 554--561, 2013.

\bibitem{Krause2013cars}
Jonathan Krause, Michael Stark, Jia Deng, and Li Fei-Fei.
\newblock 3d object representations for fine-grained categorization.
\newblock In {\em 4th International IEEE Workshop on 3D Representation and
  Recognition (3dRR-13)}, Sydney, Australia, 2013.

\bibitem{kuhl1984intermodal}
Patricia~K Kuhl and Andrew~N Meltzoff.
\newblock The intermodal representation of speech in infants.
\newblock {\em Infant behavior and development}, 7(3):361--381, 1984.

\bibitem{kumar2022fine}
Ananya Kumar, Aditi Raghunathan, Robbie Jones, Tengyu Ma, and Percy Liang.
\newblock Fine-tuning can distort pretrained features and underperform
  out-of-distribution.
\newblock {\em arXiv preprint arXiv:2202.10054}, 2022.

\bibitem{lee2019touching}
Jet-Tsyn Lee, Danushka Bollegala, and Shan Luo.
\newblock “touching to see” and “seeing to feel”: Robotic cross-modal
  sensory data generation for visual-tactile perception.
\newblock In {\em 2019 International Conference on Robotics and Automation
  (ICRA)}, pages 4276--4282. IEEE, 2019.

\bibitem{li2020caltech101}
Li, Andreeto, Ranzato, and Perona.
\newblock Caltech 101, Apr 2022.

\bibitem{li2017learning}
Ang Li, Allan Jabri, Armand Joulin, and Laurens Van Der~Maaten.
\newblock Learning visual n-grams from web data.
\newblock In {\em Proceedings of the IEEE International Conference on Computer
  Vision}, pages 4183--4192, 2017.

\bibitem{li2022blip}
Junnan Li, Dongxu Li, Caiming Xiong, and Steven Hoi.
\newblock Blip: Bootstrapping language-image pre-training for unified
  vision-language understanding and generation.
\newblock {\em arXiv preprint arXiv:2201.12086}, 2022.

\bibitem{li2021grounded}
Liunian~Harold Li*, Pengchuan Zhang*, Haotian Zhang*, Jianwei Yang, Chunyuan
  Li, Yiwu Zhong, Lijuan Wang, Lu Yuan, Lei Zhang, Jenq-Neng Hwang, Kai-Wei
  Chang, and Jianfeng Gao.
\newblock Grounded language-image pre-training.
\newblock In {\em CVPR}, 2022.

\bibitem{li2020unimo}
Wei Li, Can Gao, Guocheng Niu, Xinyan Xiao, Hao Liu, Jiachen Liu, Hua Wu, and
  Haifeng Wang.
\newblock Unimo: Towards unified-modal understanding and generation via
  cross-modal contrastive learning.
\newblock {\em arXiv preprint arXiv:2012.15409}, 2020.

\bibitem{li2021supervision}
Yangguang Li, Feng Liang, Lichen Zhao, Yufeng Cui, Wanli Ouyang, Jing Shao,
  Fengwei Yu, and Junjie Yan.
\newblock Supervision exists everywhere: A data efficient contrastive
  language-image pre-training paradigm.
\newblock {\em arXiv preprint arXiv:2110.05208}, 2021.

\bibitem{liu2021pre}
Pengfei Liu, Weizhe Yuan, Jinlan Fu, Zhengbao Jiang, Hiroaki Hayashi, and
  Graham Neubig.
\newblock Pre-train, prompt, and predict: A systematic survey of prompting
  methods in natural language processing.
\newblock {\em arXiv preprint arXiv:2107.13586}, 2021.

\bibitem{liu2021gpt}
Xiao Liu, Yanan Zheng, Zhengxiao Du, Ming Ding, Yujie Qian, Zhilin Yang, and
  Jie Tang.
\newblock Gpt understands, too.
\newblock {\em arXiv:2103.10385}, 2021.

\bibitem{lu2022prompt}
Yuning Lu, Jianzhuang Liu, Yonggang Zhang, Yajing Liu, and Xinmei Tian.
\newblock Prompt distribution learning.
\newblock In {\em Proceedings of the IEEE/CVF Conference on Computer Vision and
  Pattern Recognition}, pages 5206--5215, 2022.

\bibitem{luo2018vitac}
Shan Luo, Wenzhen Yuan, Edward Adelson, Anthony~G Cohn, and Raul Fuentes.
\newblock Vitac: Feature sharing between vision and tactile sensing for cloth
  texture recognition.
\newblock In {\em 2018 IEEE International Conference on Robotics and Automation
  (ICRA)}, pages 2722--2727. IEEE, 2018.

\bibitem{maji13aircraft}
S. Maji, J. Kannala, E. Rahtu, M. Blaschko, and A. Vedaldi.
\newblock Fine-grained visual classification of aircraft.
\newblock Technical report, 2013.

\bibitem{maji2013fine}
Subhransu Maji, Esa Rahtu, Juho Kannala, Matthew Blaschko, and Andrea Vedaldi.
\newblock Fine-grained visual classification of aircraft.
\newblock {\em arXiv preprint arXiv:1306.5151}, 2013.

\bibitem{meltzoff1979intermodal}
Andrew~N Meltzoff and Richard~W Borton.
\newblock Intermodal matching by human neonates.
\newblock {\em Nature}, 282(5737):403--404, 1979.

\bibitem{mu2019shaping}
Jesse Mu, Percy Liang, and Noah Goodman.
\newblock Shaping visual representations with language for few-shot
  classification.
\newblock {\em arXiv preprint arXiv:1911.02683}, 2019.

\bibitem{mu2022slip}
Norman Mu, Alexander Kirillov, David Wagner, and Saining Xie.
\newblock Slip: Self-supervision meets language-image pre-training.
\newblock In {\em European Conference on Computer Vision}, pages 529--544.
  Springer, 2022.

\bibitem{Nanay2018-NANMMI}
Bence Nanay.
\newblock Multimodal mental imagery.
\newblock {\em Cortex}, 105:125--136, 2018.

\bibitem{nilsback2008automated}
Maria-Elena Nilsback and Andrew Zisserman.
\newblock Automated flower classification over a large number of classes.
\newblock In {\em 2008 Sixth Indian Conference on Computer Vision, Graphics \&
  Image Processing}, pages 722--729. IEEE, 2008.

\bibitem{Nilsback08flowers}
Maria-Elena Nilsback and Andrew Zisserman.
\newblock Automated flower classification over a large number of classes.
\newblock In {\em Indian Conference on Computer Vision, Graphics and Image
  Processing}, Dec 2008.

\bibitem{pahde2018discriminative}
Frederik Pahde, Main Nabi, Tassila Klein, and Patrick Jahnichen.
\newblock Discriminative hallucination for multi-modal few-shot learning.
\newblock In {\em 2018 25th IEEE International Conference on Image Processing
  (ICIP)}, pages 156--160. IEEE, 2018.

\bibitem{pahde2021multimodal}
Frederik Pahde, Mihai Puscas, Tassilo Klein, and Moin Nabi.
\newblock Multimodal prototypical networks for few-shot learning.
\newblock In {\em Proceedings of the IEEE/CVF Winter Conference on Applications
  of Computer Vision}, pages 2644--2653, 2021.

\bibitem{parkhi2012cats}
Omkar~M Parkhi, Andrea Vedaldi, Andrew Zisserman, and CV Jawahar.
\newblock Cats and dogs.
\newblock In {\em 2012 IEEE conference on computer vision and pattern
  recognition}, pages 3498--3505. IEEE, 2012.

\bibitem{parkhi12pets}
Omkar~M. Parkhi, Andrea Vedaldi, Andrew Zisserman, and C.~V. Jawahar.
\newblock Cats and dogs.
\newblock In {\em IEEE Conference on Computer Vision and Pattern Recognition},
  2012.

\bibitem{piczak2015esc}
Karol~J Piczak.
\newblock Esc: Dataset for environmental sound classification.
\newblock In {\em Proceedings of the 23rd ACM international conference on
  Multimedia}, pages 1015--1018, 2015.

\bibitem{prasad2022grips}
Archiki Prasad, Peter Hase, Xiang Zhou, and Mohit Bansal.
\newblock Grips: Gradient-free, edit-based instruction search for prompting
  large language models.
\newblock {\em arXiv preprint arXiv:2203.07281}, 2022.

\bibitem{qi2018low}
Hang Qi, Matthew Brown, and David~G Lowe.
\newblock Low-shot learning with imprinted weights.
\newblock In {\em Proceedings of the IEEE conference on computer vision and
  pattern recognition}, pages 5822--5830, 2018.

\bibitem{quiroga2005invariant}
R~Quian Quiroga, Leila Reddy, Gabriel Kreiman, Christof Koch, and Itzhak Fried.
\newblock Invariant visual representation by single neurons in the human brain.
\newblock {\em Nature}, 435(7045):1102--1107, 2005.

\bibitem{radford2021learning}
Alec Radford, Jong~Wook Kim, Chris Hallacy, Aditya Ramesh, Gabriel Goh,
  Sandhini Agarwal, Girish Sastry, Amanda Askell, Pamela Mishkin, Jack Clark,
  et~al.
\newblock Learning transferable visual models from natural language
  supervision.
\newblock In {\em ICML}. PMLR, 2021.

\bibitem{ravi2016optimization}
Sachin Ravi and Hugo Larochelle.
\newblock Optimization as a model for few-shot learning.
\newblock 2016.

\bibitem{recht2019imagenet}
Benjamin Recht, Rebecca Roelofs, Ludwig Schmidt, and Vaishaal Shankar.
\newblock Do imagenet classifiers generalize to imagenet?
\newblock In {\em International Conference on Machine Learning}, pages
  5389--5400. PMLR, 2019.

\bibitem{schick2020exploiting}
Timo Schick and Hinrich Schütze.
\newblock Exploiting cloze questions for few-shot text classification and
  natural language inference.
\newblock {\em Computing Research Repository}, arXiv:2001.07676, 2020.

\bibitem{schick2020small}
Timo Schick and Hinrich Schütze.
\newblock It's not just size that matters: Small language models are also
  few-shot learners.
\newblock {\em Computing Research Repository}, arXiv:2009.07118, 2020.

\bibitem{schmidt2009meaning}
Lauren~A Schmidt.
\newblock {\em Meaning and compositionality as statistical induction of
  categories and constraints}.
\newblock PhD thesis, Massachusetts Institute of Technology, 2009.

\bibitem{scholkopf2001generalized}
Bernhard Sch{\"o}lkopf, Ralf Herbrich, and Alex~J Smola.
\newblock A generalized representer theorem.
\newblock In {\em International conference on computational learning theory},
  pages 416--426. Springer, 2001.

\bibitem{schwartz2022baby}
Eli Schwartz, Leonid Karlinsky, Rogerio Feris, Raja Giryes, and Alex Bronstein.
\newblock Baby steps towards few-shot learning with multiple semantics.
\newblock {\em Pattern Recognition Letters}, 160:142--147, 2022.

\bibitem{shin2020autoprompt}
Taylor Shin, Yasaman Razeghi, Robert~L Logan~IV, Eric Wallace, and Sameer
  Singh.
\newblock Autoprompt: Eliciting knowledge from language models with
  automatically generated prompts.
\newblock {\em arXiv preprint arXiv:2010.15980}, 2020.

\bibitem{smith2005development}
Linda Smith and Michael Gasser.
\newblock The development of embodied cognition: Six lessons from babies.
\newblock {\em Artificial life}, 11(1-2):13--29, 2005.

\bibitem{snell2017prototypical}
Jake Snell, Kevin Swersky, and Richard Zemel.
\newblock Prototypical networks for few-shot learning.
\newblock {\em Advances in neural information processing systems}, 30, 2017.

\bibitem{song2022clip}
Haoyu Song, Li Dong, Wei-Nan Zhang, Ting Liu, and Furu Wei.
\newblock Clip models are few-shot learners: Empirical studies on vqa and
  visual entailment.
\newblock {\em arXiv preprint arXiv:2203.07190}, 2022.

\bibitem{soomro2012ucf101}
Khurram Soomro, Amir~Roshan Zamir, and Mubarak Shah.
\newblock Ucf101: A dataset of 101 human actions classes from videos in the
  wild.
\newblock {\em arXiv preprint arXiv:1212.0402}, 2012.

\bibitem{tsimpoukelli2021multimodal}
Maria Tsimpoukelli, Jacob~L Menick, Serkan Cabi, SM Eslami, Oriol Vinyals, and
  Felix Hill.
\newblock Multimodal few-shot learning with frozen language models.
\newblock {\em Advances in Neural Information Processing Systems}, 34:200--212,
  2021.

\bibitem{NIPS2016_90e13578}
Oriol Vinyals, Charles Blundell, Timothy Lillicrap, koray kavukcuoglu, and Daan
  Wierstra.
\newblock Matching networks for one shot learning.
\newblock In D. Lee, M. Sugiyama, U. Luxburg, I. Guyon, and R. Garnett,
  editors, {\em Advances in Neural Information Processing Systems}, volume~29.
  Curran Associates, Inc., 2016.

\bibitem{wang2019learning}
Haohan Wang, Songwei Ge, Zachary Lipton, and Eric~P Xing.
\newblock Learning robust global representations by penalizing local predictive
  power.
\newblock In {\em Advances in Neural Information Processing Systems}, pages
  10506--10518, 2019.

\bibitem{wang2022debiased}
Xudong Wang, Zhirong Wu, Long Lian, and Stella~X Yu.
\newblock Debiased learning from naturally imbalanced pseudo-labels.
\newblock In {\em Proceedings of the IEEE/CVF Conference on Computer Vision and
  Pattern Recognition}, pages 14647--14657, 2022.

\bibitem{wang2018low}
Yu-Xiong Wang, Ross Girshick, Martial Hebert, and Bharath Hariharan.
\newblock Low-shot learning from imaginary data.
\newblock In {\em Proceedings of the IEEE conference on computer vision and
  pattern recognition}, pages 7278--7286, 2018.

\bibitem{wang2017growing}
Yu-Xiong Wang, Deva Ramanan, and Martial Hebert.
\newblock Growing a brain: Fine-tuning by increasing model capacity.
\newblock In {\em Proceedings of the IEEE Conference on Computer Vision and
  Pattern Recognition}, pages 2471--2480, 2017.

\bibitem{wortsman2022robust}
Mitchell Wortsman, Gabriel Ilharco, Jong~Wook Kim, Mike Li, Simon Kornblith,
  Rebecca Roelofs, Raphael~Gontijo Lopes, Hannaneh Hajishirzi, Ali Farhadi,
  Hongseok Namkoong, et~al.
\newblock Robust fine-tuning of zero-shot models.
\newblock In {\em Proceedings of the IEEE/CVF Conference on Computer Vision and
  Pattern Recognition}, pages 7959--7971, 2022.

\bibitem{wu2022transferring}
Wenhao Wu, Zhun Sun, and Wanli Ouyang.
\newblock Transferring textual knowledge for visual recognition.
\newblock {\em arXiv preprint arXiv:2207.01297}, 2022.

\bibitem{xian2018zero}
Yongqin Xian, Christoph~H Lampert, Bernt Schiele, and Zeynep Akata.
\newblock Zero-shot learning—a comprehensive evaluation of the good, the bad
  and the ugly.
\newblock {\em IEEE transactions on pattern analysis and machine intelligence},
  41(9):2251--2265, 2018.

\bibitem{xiao2010sun}
Jianxiong Xiao, James Hays, Krista~A Ehinger, Aude Oliva, and Antonio Torralba.
\newblock Sun database: Large-scale scene recognition from abbey to zoo.
\newblock In {\em 2010 IEEE computer society conference on computer vision and
  pattern recognition}, pages 3485--3492. IEEE, 2010.

\bibitem{xing2019adaptive}
Chen Xing, Negar Rostamzadeh, Boris Oreshkin, and Pedro~O O~Pinheiro.
\newblock Adaptive cross-modal few-shot learning.
\newblock {\em Advances in Neural Information Processing Systems}, 32, 2019.

\bibitem{xing2022class}
Yinghui Xing, Qirui Wu, De Cheng, Shizhou Zhang, Guoqiang Liang, and Yanning
  Zhang.
\newblock Class-aware visual prompt tuning for vision-language pre-trained
  model.
\newblock {\em arXiv preprint arXiv:2208.08340}, 2022.

\bibitem{yu2022coca}
Jiahui Yu, Zirui Wang, Vijay Vasudevan, Legg Yeung, Mojtaba Seyedhosseini, and
  Yonghui Wu.
\newblock Coca: Contrastive captioners are image-text foundation models.
\newblock {\em arXiv preprint arXiv:2205.01917}, 2022.

\bibitem{zhai2022lit}
Xiaohua Zhai, Xiao Wang, Basil Mustafa, Andreas Steiner, Daniel Keysers,
  Alexander Kolesnikov, and Lucas Beyer.
\newblock Lit: Zero-shot transfer with locked-image text tuning.
\newblock In {\em Proceedings of the IEEE/CVF Conference on Computer Vision and
  Pattern Recognition}, pages 18123--18133, 2022.

\bibitem{zhang2021cross}
Han Zhang, Jing~Yu Koh, Jason Baldridge, Honglak Lee, and Yinfei Yang.
\newblock Cross-modal contrastive learning for text-to-image generation.
\newblock In {\em Proceedings of the IEEE/CVF conference on computer vision and
  pattern recognition}, pages 833--842, 2021.

\bibitem{zhang2022glipv2}
Haotian* Zhang, Pengchuan* Zhang, Xiaowei Hu, Yen-Chun Chen, Liunian~Harold Li,
  Xiyang Dai, Lijuan Wang, Lu Yuan, Jenq-Neng Hwang, and Jianfeng Gao.
\newblock Glipv2: Unifying localization and vision-language understanding.
\newblock {\em arXiv preprint arXiv:2206.05836}, 2022.

\bibitem{zhang2020side}
Jeffrey~O Zhang, Alexander Sax, Amir Zamir, Leonidas Guibas, and Jitendra
  Malik.
\newblock Side-tuning: a baseline for network adaptation via additive side
  networks.
\newblock In {\em European Conference on Computer Vision}, pages 698--714.
  Springer, 2020.

\bibitem{zhang2021tip}
Renrui Zhang, Rongyao Fang, Peng Gao, Wei Zhang, Kunchang Li, Jifeng Dai, Yu
  Qiao, and Hongsheng Li.
\newblock Tip-adapter: Training-free clip-adapter for better vision-language
  modeling.
\newblock {\em arXiv preprint arXiv:2111.03930}, 2021.

\bibitem{zhou2022cocoop}
Kaiyang Zhou, Jingkang Yang, Chen~Change Loy, and Ziwei Liu.
\newblock Conditional prompt learning for vision-language models.
\newblock In {\em CVPR}, 2022.

\bibitem{zhou2022coop}
Kaiyang Zhou, Jingkang Yang, Chen~Change Loy, and Ziwei Liu.
\newblock Learning to prompt for vision-language models.
\newblock {\em IJCV}, 2022.

\bibitem{zhu2022prompt}
Beier Zhu, Yulei Niu, Yucheng Han, Yue Wu, and Hanwang Zhang.
\newblock Prompt-aligned gradient for prompt tuning.
\newblock {\em arXiv preprint arXiv:2205.14865}, 2022.

\end{thebibliography}
}

\clearpage

\section*{}
\begin{center}
    {\bf \large Appendix}
\end{center}

\section{Experimental Details}\label{sec:hyperparameters}
In this section, we go through the hyperparameter details for all the experiments for reproducibility. 

{\bf Basic settings:} We follow the original CLIP~\cite{radford2021learning} to L2-normalize the features after the encoder before sending them into the linear layer. We also use the L2-normalized text features to initialize the final linear layer weight following WiSE-FT~\cite{wortsman2022robust}. For all cross-modal adaptation experiments, half of the batch is image samples and the other half is text samples. For all experiments, we use AdamW optimizer following WiSE-FT~\cite{wortsman2022robust} and tune the hyperparameters including initial learning rate, weight decay, and batch size on the few-shot validation set. We perform a learning rate warmup with 50 iterations, during which the learning rate goes up linearly from 0.00001 to the initial value. We then perform a cosine annealing learning rate scheduling over the course of 12800 iterations. We do early stopping based on the few-shot validation set performance evaluated every 100 iterations. Furthermore, because the logit scale (inverse of softmax temperature) is a learnable weight clipped at 100 during CLIP-pretraining~\cite{radford2021learning}, we reuse the given logit scale of 100 for all experiments except for partial finetuning, where we find lowering it to 50 can improve validation performance. Future work may choose to set the logit scale as a learnable parameter instead.

We now report the range of hyperparameter search for each method. Note that the search range is kept the same for all 11 target datasets.

{\bf Linear Probing:} For all linear probing experiments, we perform a grid search of learning rate in $[0.001, 0.0001]$, weight decay in $[0.0, 0.01, 0.0001]$, and batch size in $[8, 32]$. 

{\bf WiSE-FT:} To compare with linear probing, we adopt the same procedure above to train the linear classifier and then perform post-hoc ensembling with the text-based classifier with a fixed ratio of 0.5.

{\bf Partial Finetuning:} For all partial finetuning experiments, we perform a grid search of learning rate in $[0.00001,0.000001, 0.0000001]$, weight decay in $[0.0, 0.001, 0.00001]$, and batch size is set to 8. CLIP~\cite{radford2021learning} adopts a modified version of ResNet-50 image encoder, in which the final average pooling layer is replaced by an attentional pooling layer. We thus choose this layer as the finetuning target for all ResNet-50 experiments. For ViT-B/16 encoder, we simply finetune the last transformer layer. In the next section, we also show that finetuning the text encoder is not as effective.

{\bf Cross-modal Prompting:} We follow the same setup and hyperparameters used in CoOp~\cite{zhou2022coop}. We use the ResNet-50 backbone with 16 learnable tokens, and append the class name to the end of the tokens. Following CoOp, we use SGD with a learning rate of $0.002$, decayed using the cosine annealing rule. We train for 200 epochs for 8 and 16 shots, 100 epochs for 2 and 4 shots, and 50 epochs for 1 shot (except ImageNet which is fixed at 50 epochs). The learning rate for the first epoch is fixed at $0.00001$. We also use the same random resized crop transformations as CoOp. 

{\bf Cross-modal Adapter:} We follow the same 2-layer MLPs architecture in CLIP-Adapter~\cite{gao2021clip} with a residual ratio of $0.2$. Specifically, the first linear layer downsizes the input feature to $\frac{1}{4}$ of the original dimension and the second linear layer transforms it back to the original dimension. Each linear layer is followed by a ReLU function. Finally, the transformed features are multiplied by 0.2 and added with 0.8 * the original feature. We use a single adapter for both image and text features. We perform a grid search of learning rate in $[0.0001, 0.00001, 0.000001, 0.0000001]$, weight decay in $[0.0, 0.001, 0.00001]$, and batch size is set to 8. We do not adopt the cache-modal and training-free initialization proposed in the follow-up Tip-Adapter~\cite{zhang2021tip} method. Also, we notice that Tip-Adapter uses test set to perform early stopping; we however strictly follow the CoOp protocol to use the few-shot validation set for all hyperparameter searching.

{\bf ImageNet-ESC Experiments:} For all linear probing experiments on ImageNet-ESC, we perform a grid search of learning rate in $[0.1, 0.01, 0.001, 0.0001]$, weight decay in $[0.0, 0.01, 0.0001]$, and batch size is 8. 

\begin{table}[t]
\centering
\resizebox{\linewidth}{!}{
\tiny
\begin{tabular}{cc|c}
\toprule 
Included Dataset & ESC-50~\cite{piczak2015esc} Class & ImageNet~\cite{deng2009imagenet} Class \\
\midrule
\multirow{19}{*}{ImageNet-ESC-19} & {\tt rooster} & {\tt rooster} \\
& {\tt hen} & {\tt hen} \\
& {\tt chirping-birds} & {\tt chickadee} \\
& {\tt frog} & {\tt tree frog} \\
& {\tt dog} & {\tt otterhound} \\
& {\tt cat} & {\tt egyptian cat} \\
& {\tt insects} & {\tt fly} \\
& {\tt crickets} & {\tt cricket} \\
& {\tt pig} & {\tt pig} \\
& {\tt sheep} & {\tt big-horn sheep} \\
& {\tt airplane} & {\tt airliner} \\
& {\tt train} & {\tt high-speed train} \\
& {\tt chainsaw} & {\tt chainsaw} \\
& {\tt keyboard-typing} & {\tt computer keyboard} \\
& {\tt clock-alarm} & {\tt digital clock} \\
& {\tt mouse-click} & {\tt computer mouse} \\
& {\tt vacuum-cleaner} & {\tt vacuum cleaner} \\
& {\tt clock-tick} & {\tt wall clock} \\
& {\tt washing-machine} & {\tt washing machine} \\
\hline
\multirow{8}{*}{ImageNet-ESC-27} & {\tt can-opening} & {\tt can opener} \\
& {\tt church-bells} & {\tt church bells} \\
& {\tt crackling-fire} & {\tt fire screen} \\
& {\tt toilet-flush} & {\tt toilet seat} \\
& {\tt water-drops} & {\tt sink} \\
& {\tt drinking-sipping} & {\tt water bottle} \\
& {\tt pouring-water} & {\tt water jug} \\
& {\tt sea-waves} & {\tt sandbar} \\
    \bottomrule
    \end{tabular}
}
\caption{\small \textbf{ImageNet-ESC dataset class matchings.}}
\vspace{-4mm}
\label{tab:imagenet_esc_dataset}
\end{table}

\clearpage
\section{Additional Results}\label{sec:all_results}

\begin{figure*}
    \centering
    \begin{subfigure}{0.33 \textwidth}
        \includegraphics[width=\textwidth]{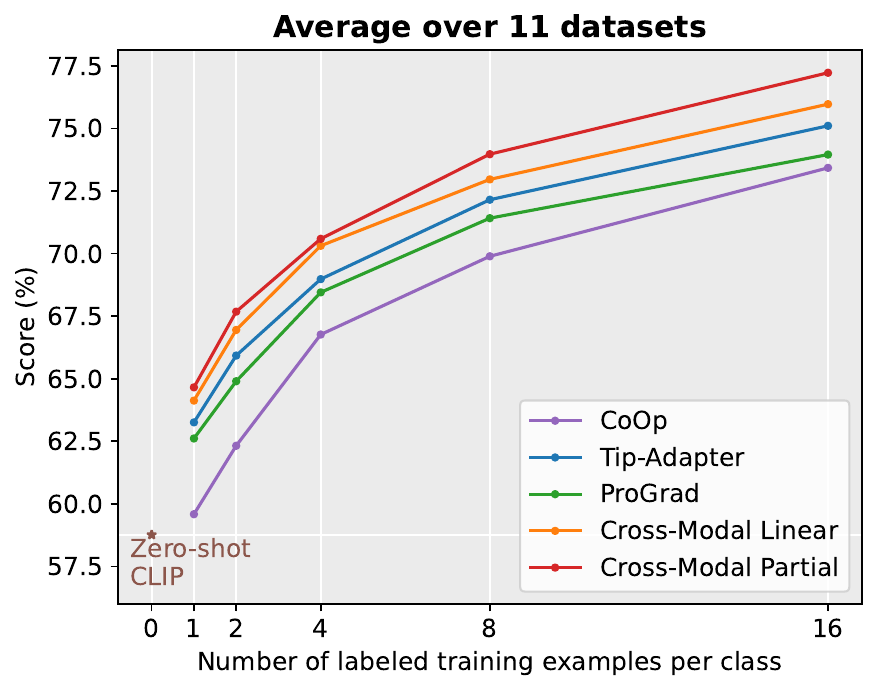}
    \end{subfigure} \hfill
     \begin{subfigure}{0.33 \textwidth}
        \includegraphics[width=\textwidth]{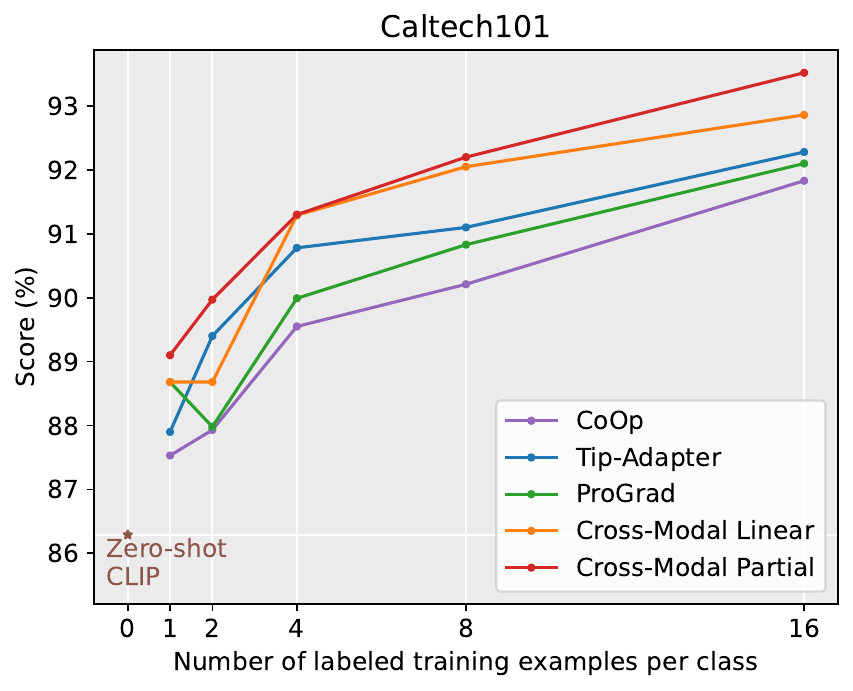}
    \end{subfigure} \hfill
     \begin{subfigure}{0.33 \textwidth}
        \includegraphics[width=\textwidth]{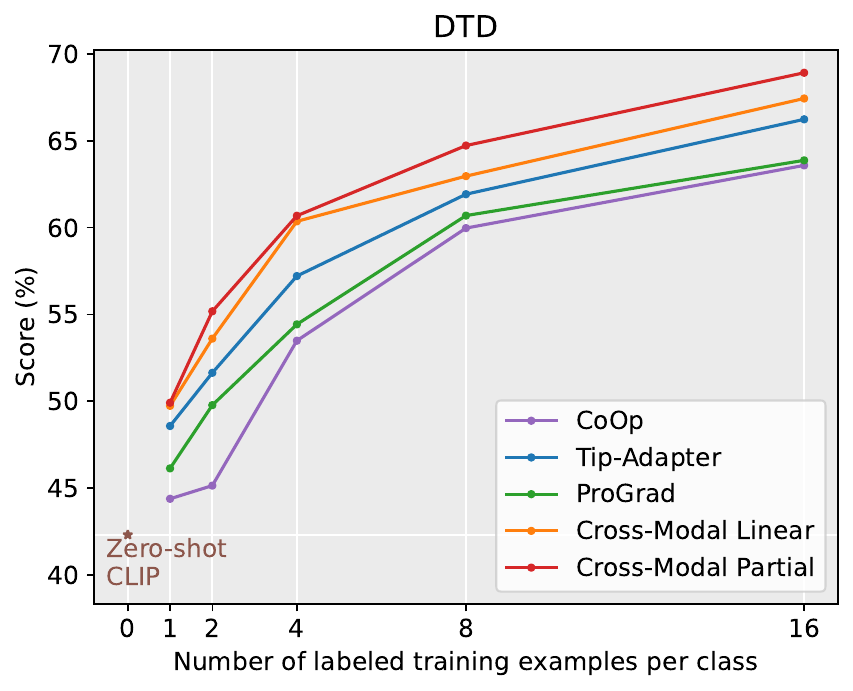}
    \end{subfigure}
     \begin{subfigure}{0.33 \textwidth}
        \includegraphics[width=\textwidth]{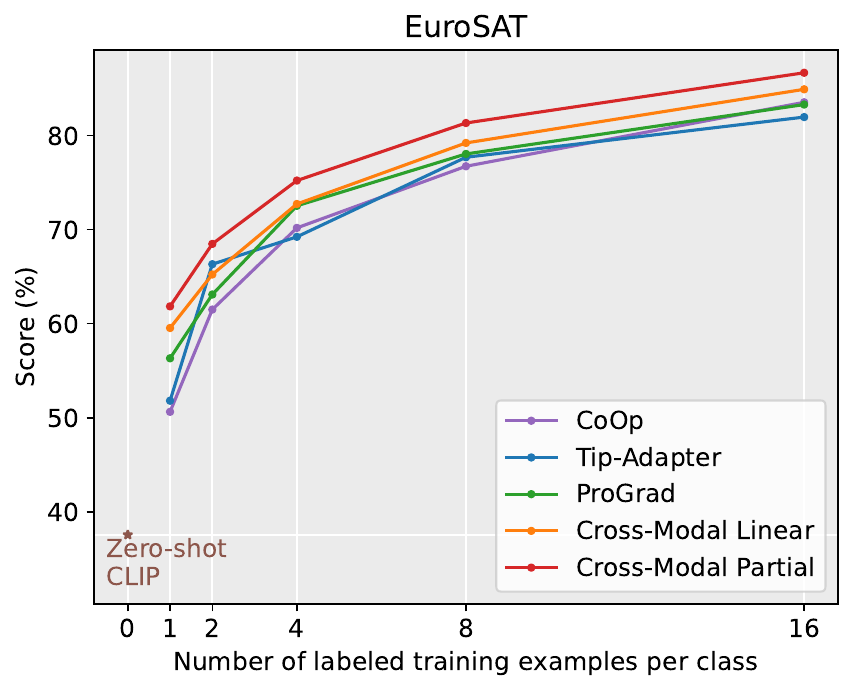}
    \end{subfigure}
     \begin{subfigure}{0.33 \textwidth}
        \includegraphics[width=\textwidth]{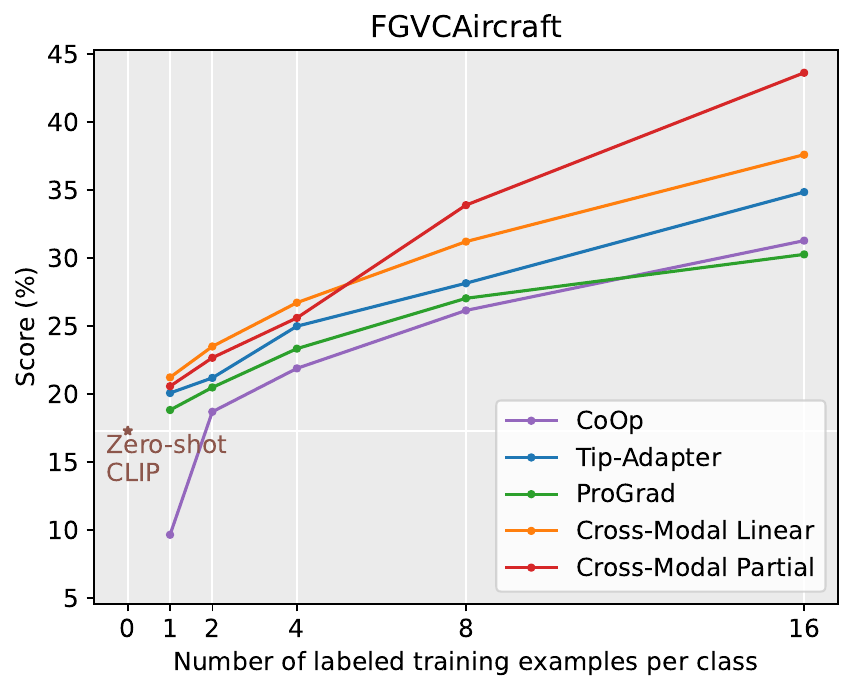}
    \end{subfigure}
     \begin{subfigure}{0.33 \textwidth}
        \includegraphics[width=\textwidth]{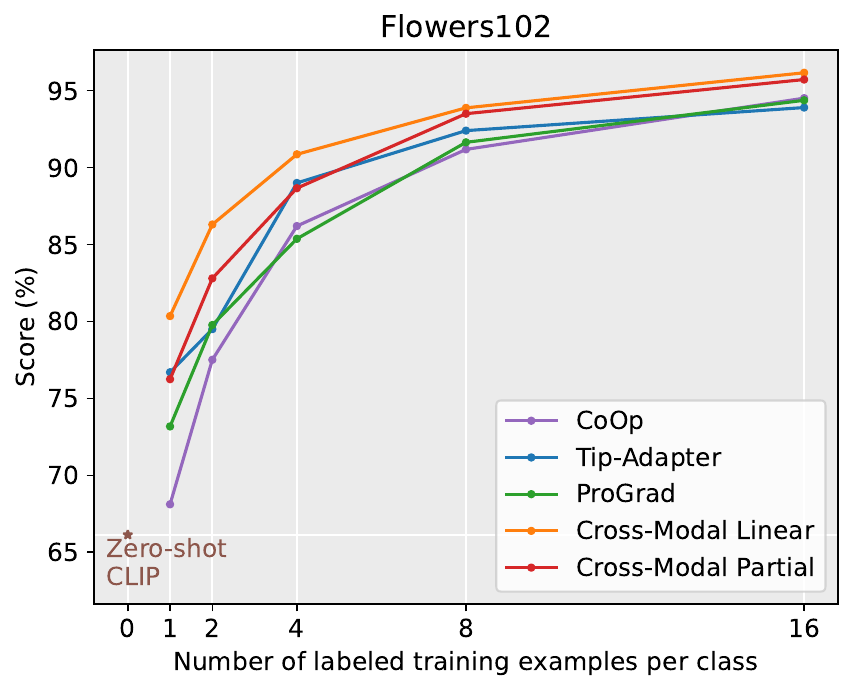}
    \end{subfigure}
     \begin{subfigure}{0.33 \textwidth}
        \includegraphics[width=\textwidth]{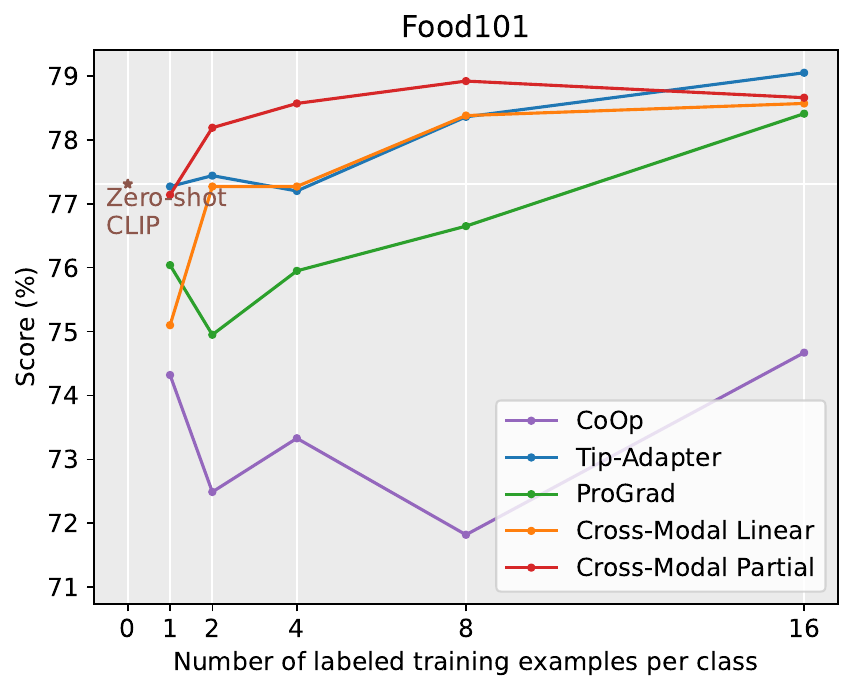}
    \end{subfigure}
     \begin{subfigure}{0.33 \textwidth}
        \includegraphics[width=\textwidth]{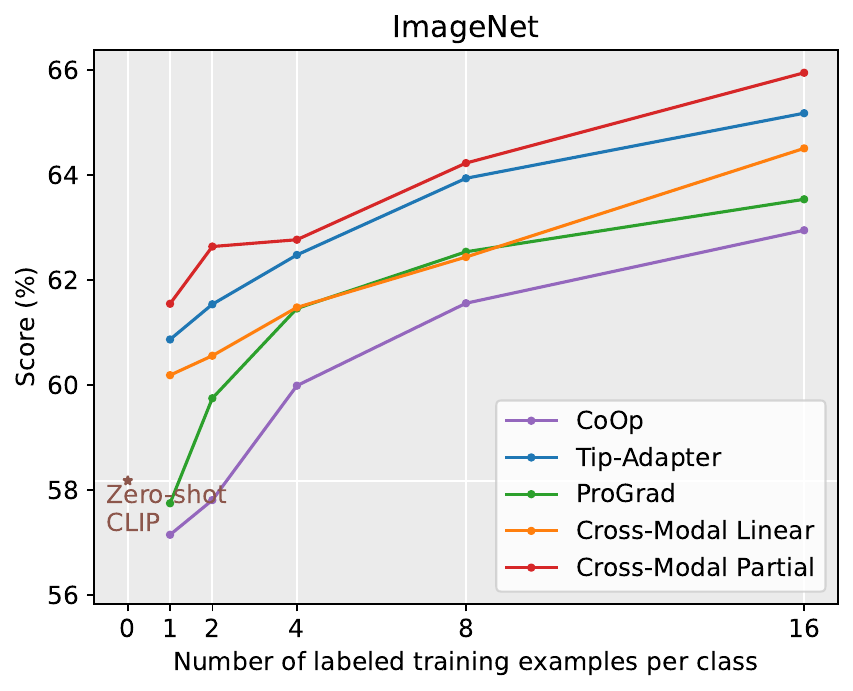}
    \end{subfigure}
     \begin{subfigure}{0.33 \textwidth}
        \includegraphics[width=\textwidth]{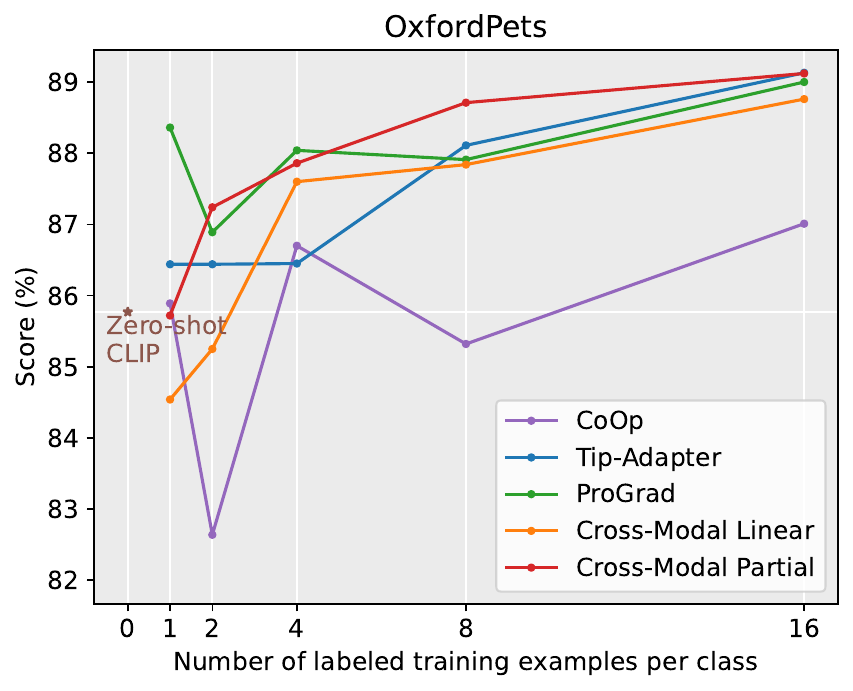}
    \end{subfigure} \begin{subfigure}{0.33 \textwidth}
        \includegraphics[width=\textwidth]{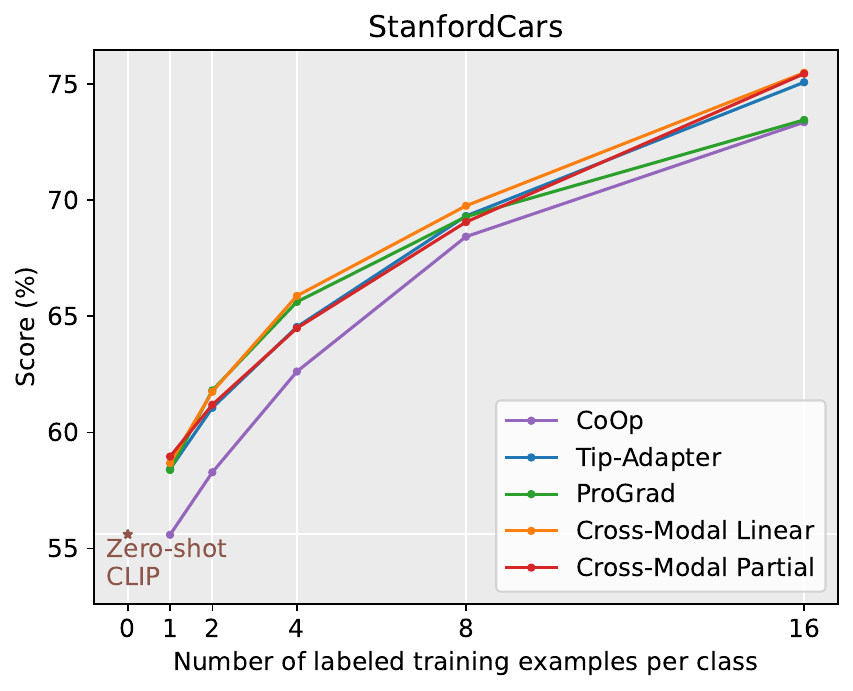}
    \end{subfigure}
     \begin{subfigure}{0.33 \textwidth}
        \includegraphics[width=\textwidth]{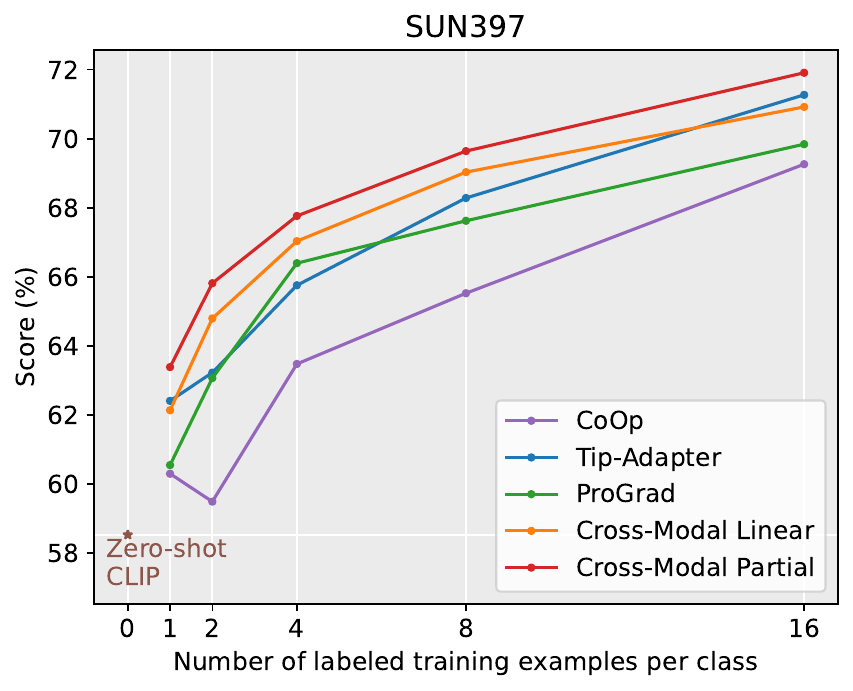}
    \end{subfigure}
     \begin{subfigure}{0.33 \textwidth}
        \includegraphics[width=\textwidth]{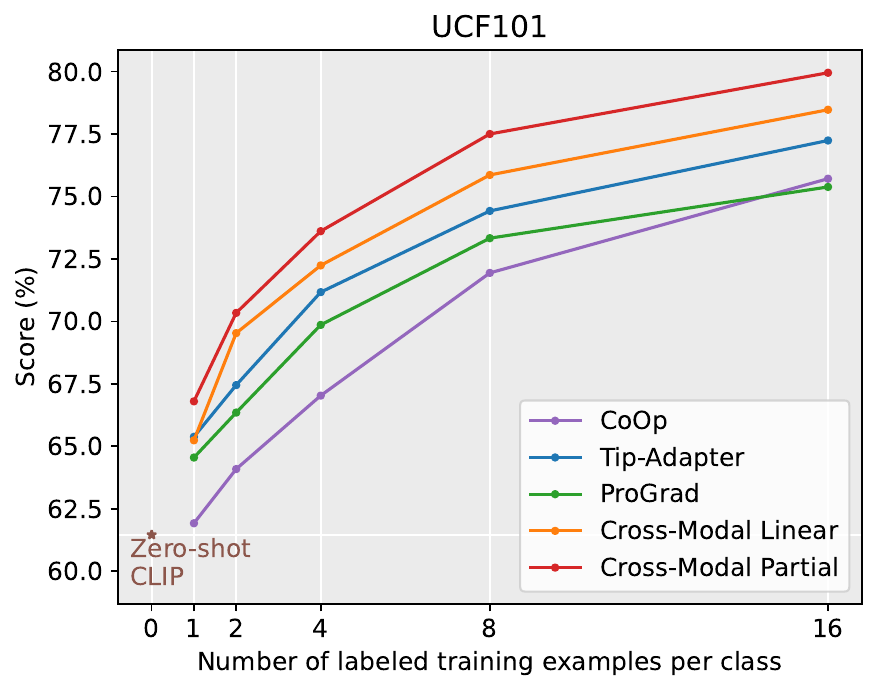}
    \end{subfigure}
    \caption{\textbf{Comparison of few-shot learning results across 11 datasets.} We show our main methods (cross-modal linear probing and partial finetuning) and compare them with prior works. We note that the Tip-Adapter~\cite{zhang2021tip} numbers shown are \underline{our own re-run} of the method, where we replace their early-stopping on the test set with early stopping on the few-shot validation set for a fair comparison. As seen in the plots, cross-modal partial finetuning consistently outperforms prior works across the datasets, and cross-modal linear probing is also generally more performant.  }
    \label{fig:per_dataset}
\end{figure*}

In this section, we present all the results with standard deviation over multiple runs. Here is an overview (please refer to table captions for more discussion):
\begin{enumerate}
    \item {\bf Per-dataset results for all methods:} We show \autoref{fig:per_dataset} and \autoref{tab:per_dataset}. In particular, we note that cross-modal adaptation consistently outperforms prior methods across a wide variety of visual recognition datasets, further strengthening our claim that our approach should be the de-facto adaptation method for finetuning multimodal models.
    \item {\bf Ablation for augmentation techniques:} In \autoref{tab:augmentation_all}, we show the performance of all combinations of image and text augmentation techniques. Importantly, simple {\em text} augmentation strategies work very well for {\em visual} recognition. 
    \item {\bf Ablation for classifier initialization:} In \autoref{tab:init_results}, our experiments suggest that (a) text-based initialization is beneficial for both linear and partial finetuning, and (b) cross-modal adaptation can improve the performance regardless of the initialization.
    \item {\bf Ablation for partial finetuning:} In \autoref{tab:partial_results}, we confirm that partial finetuning of the image encoder is more effective than finetuning the text encoder.
    \item {\bf Complete results for all reported methods:} In \autoref{tab:complete_results}, we show the standard deviation for all methods reported in the main paper and appendix, including ViT-based encoder results. 
    \item {\bf Complete results on ImageNet-ESC benchmark:} We show the complete results on ImageNet-ESC-19 and ImageNet-ESC-27 for both image-classification in \autoref{tab:image_complete} and audio-classification in \autoref{tab:audio_complete}. We additionally include the results of the text-based classifier and cross-modal linear probing with all three modalities (including text) for reference. Including the text modality seems to be the most performant, which is expected since the benchmark is curated based on textual information, i.e. matching label names. We also note that just adding text modality is better than including all three modalities; we believe this issue can be alleviated with better alignment between the image and audio representations, e.g. scaling the pre-training data for AudioCLIP. Furthermore, the standard deviations of the experiments are higher than those of the vision-language adaptation experiments because the randomly sampled one-shot sample can make a huge difference in the performance. However, cross-modal adaptation is more performant not by chance -- in more than $75\%$ of the experiments, adding the one-shot-audio or one-shot-image to the same set of samples can outperform uni-modal linear probing.
    \item {\bf Comparison to ProDA~\cite{lu2022prompt}:} In \autoref{tab:comparison_to_proda}, we compare to ProDA, another promising SOTA method that does automatic prompt ensembling with 36 learned templates. We are told by the authors that they do not follow the dataset split given by CoOp~\cite{zhou2022coop}, and use the official test split of each dataset whenever possible or sample their own test split from the train set. Therefore, we cannot directly compare to their performance since CoOp~\cite{zhou2022coop} use their own test split for most datasets and ProDA does not release the code yet. In particular, official test sets exist for two of the target datasets (Food101~\cite{bossard2014food} and DTD~\cite{cimpoi2014describing}). We therefore switch to the official test split for these two datasets and use the CoOp's split for the rest of the 9 datasets in \autoref{tab:comparison_to_proda} as our best attempt to compare to ProDA~\cite{lu2022prompt}. Note that ProDa also does not report the use of a few-shot validation set. In conclusion, our approach is still more performant than theirs under most scenarios with significantly fewer training resources.
    \item {\bf 180 templates used for mining:} In \autoref{tab:180_templates}, we show the pool of templates we use when mining based on few-shot validation set performance. 
\end{enumerate}

\clearpage
\begin{table*}[h!]
    \centering
    \renewcommand{\arraystretch}{1.2}
    \resizebox{\textwidth}{!}{
    \begin{tabular}{c|c|ccccccccccc|c}
    \toprule
       \multirow{2}{*}{\textbf{Method}}  & \multirow{2}{*}{\textbf{Shots}}  &  \multicolumn{11}{c}{\textbf{Dataset}}\\ \cmidrule(l){3-14} 
         &  & Caltech~\cite{li2020caltech101} & ImageNet~\cite{deng2009imagenet} & DTD~\cite{cimpoi14dtd} & EuroSAT~\cite{helber2017eurosat} & Aircraft~\cite{maji13aircraft} & Food~\cite{bossard14food} & Flowers~\cite{Nilsback08flowers} & Pets~\cite{parkhi12pets} & Cars~\cite{Krause2013cars} & SUN397~\cite{xiao2010sun} & UCF101~\cite{soomro2012ucf101} & Average \\ 

    \midrule 
    Zero-Shot CLIP  & 0     & $86.29$          & $58.18$          & $42.32$          & $37.56$          & $17.28$          & $77.31$          & $66.14$          & $85.77$          & $55.61$          & $58.52$          & $61.46$          & $58.77$          \\ \hline
\multirow{5}{*}{CoOp}                & 1     & $87.53$          & $57.15$          & $44.39$          & $50.63$          & $9.64$           & $74.32$          & $68.12$          & $85.89$          & $55.59$          & $60.29$          & $61.92$          & $59.77$          \\
                    & 2     & $87.93$          & $57.81$          & $45.15$          & $61.50$          & $18.68$          & $72.49$          & $77.51$          & $82.64$          & $58.28$          & $59.48$          & $64.09$          & $62.32$          \\ 
                    & 4     & $89.55$          & $59.99$          & $53.49$          & $70.18$          & $21.87$          & $73.33$          & $86.20$          & $86.70$          & $62.62$          & $63.47$          & $67.03$          & $66.77$          \\
                    & 8     & $90.21$          & $61.56$          & $59.97$          & $76.73$          & $26.13$          & $71.82$          & $91.18$          & $85.32$          & $68.43$          & $65.52$          & $71.94$          & $69.89$          \\
                    & 16    & $91.83$          & $62.95$          & $63.58$          & $83.53$          & $31.26$          & $74.67$          & $94.51$          & $87.01$          & $73.36$          & $69.26$          & $75.71$          & $73.42$          \\ \midrule 
\multirow{5}{*}{Tip-Adapter}         & 1     & $87.90 \pm 0.75$ & $60.87 \pm 0.04$  & $48.58 \pm 0.63$    & $51.81 \pm 2.45$    & $20.06 \pm 0.39$ & $\mathbf{77.27 \pm 0.39}$   & $76.70 \pm 0.28$          & $86.44 \pm 1.35$       & $58.42 \pm 0.47$          & $62.40 \pm 0.27 $          & $65.38 \pm 0.29$          & $63.26 \pm 0.68$          \\
                    & 2     & \underline{$89.40 \pm 0.22$ }         & $61.54 \pm 0.01$          & $51.64 \pm 0.58$          & \underline{$66.32 \pm 2.06$}        & $21.17 \pm 0.62$       & $77.44 \pm 0.07$       & $79.50 \pm 1.07$          & {$86.44 \pm 0.44$}        & $61.06 \pm 0.41$          & $63.22 \pm 0.62$          & $67.45 \pm 1.77$          & $65.93 \pm 0.72$          \\
                    & 4     & $90.78 \pm 0.16$          & $62.48 \pm 0.01$          & $57.21 \pm 0.33$          & $69.23 \pm 2.85$        & $24.97 \pm 0.84$          & $77.20 \pm 0.43$          & $89.00 \pm 0.44$          & $86.45 \pm 0.71$          & $64.54 \pm 0.38$          & $65.75 \pm 0.15$          & $71.17 \pm 0.36$          & $68.98 \pm 0.61$          \\
                    & 8     & $91.10 \pm 0.18$          & \underline{$63.94 \pm 0.16$}       & $61.92 \pm 0.83$          & $77.69 \pm 2.45$          & $28.13 \pm 1.06$          & $78.36 \pm 0.12$          & $92.40 \pm 0.24$          & $88.11 \pm 0.42$          & $69.32 \pm 0.08$          & $68.28 \pm 0.34$          & $74.42 \pm 0.72$          & $72.15 \pm 0.60$          \\
                    & 16    & $92.28 \pm 0.66$          & \underline{$65.18 \pm 0.15$}          & $66.23 \pm 0.79$          & $81.96 \pm 2.26$          & $34.83 \pm 0.92$          & $\underline{79.05 \pm 0.26}$          & $93.90 \pm 0.68$          & $\underline{89.13 \pm 0.28}$          & $75.08 \pm 0.23$          & $71.27 \pm 0.13$          & $77.24 \pm 0.3$          & $75.10 \pm 0.61$          \\ \midrule 
\multirow{5}{*}{ProGrad}             & 1     & $88.68 \pm 0.34$ & $57.75 \pm 0.24$ & $46.14 \pm 1.74$ & $56.32 \pm 3.04$ & $18.81 \pm 0.50$ & $76.04 \pm 0.54$ & $73.18 \pm 0.73$ & $\mathbf{88.36 \pm 0.73}$ & $58.38 \pm 0.23$ & $60.54 \pm 0.24$ & $64.55 \pm 0.50$ & $62.61 \pm 0.80$ \\
                    & 2     & $87.98 \pm 0.69$ & $59.75 \pm 0.33$ & $49.78 \pm 1.37$ & $63.10 \pm 3.77$ & $20.47 \pm 0.90$ & $74.95 \pm 0.57$ & $79.77 \pm 0.65$ & $86.89 \pm 0.42$ & $61.81 \pm 0.45$ & $63.06 \pm 0.11$ & $66.35 \pm 0.18$ & $64.90 \pm 0.86$ \\
                    & 4     & $89.99 \pm 0.26$ & $61.46 \pm 0.07$ & $54.43 \pm 0.86$ & $72.53 \pm 1.29$ & $23.32 \pm 0.36$ & $75.95 \pm 0.27$ & $85.37 \pm 0.96$ & $\mathbf{88.04 \pm 0.50}$ & $65.62 \pm 0.43$ & $66.39 \pm 0.43$ & $69.86 \pm 0.30$ & $68.45 \pm 0.52$ \\
                    & 8     & $90.83 \pm 0.07$ & $62.54 \pm 0.03$ & $60.69 \pm 0.10$ & $78.04 \pm 2.45$ & $27.02 \pm 0.67$ & $76.65 \pm 0.23$ & $91.64 \pm 0.24$ & $87.91 \pm 0.54$ & $69.29 \pm 0.11$ & $67.62 \pm 0.28$ & $73.33 \pm 0.65$ & $71.41 \pm 0.49$ \\
                    & 16    & $92.10 \pm 0.39$ & $63.54 \pm 0.08$ & $63.87 \pm 0.99$ & $83.29 \pm 0.85$ & $30.25 \pm 1.09$ & $78.41 \pm 0.08$ & $94.37 \pm 0.24$ & $89.00 \pm 0.32$ & $73.46 \pm 0.29$ & $69.84 \pm 0.18$ & $75.38 \pm 0.10$ & $73.96 \pm 0.42$ \\ \midrule 
\multirow{5}{*}{Wise-FT}             & 1     & $85.49 \pm 0.81$ & $58.30 \pm 0.24$ & $44.17 \pm 0.72$ & $52.30 \pm 2.00$ & $18.61 \pm 0.54$ & $71.88 \pm 0.02$ & $65.83 \pm 0.54$ & $81.73 \pm 1.15$ & $55.66 \pm 0.15$ & $56.59 \pm 0.10$ & $59.39 \pm 1.33$ & $59.09 \pm 0.69$ \\
                    & 2     & $87.00 \pm 0.68$ & $59.08 \pm 0.34$ & $46.95 \pm 0.27$ & $57.07 \pm 4.26$ & $20.88 \pm 0.36$ & $73.54 \pm 0.11$ & $71.02 \pm 0.94$ & $82.75 \pm 0.62$ & $58.67 \pm 0.15$ & $60.15 \pm 0.10$ & $62.74 \pm 0.67$ & $61.80 \pm 0.77$ \\
                    & 4     & $89.03 \pm 0.17$ & $60.48 \pm 0.11$ & $52.23 \pm 0.70$ & $62.45 \pm 4.09$ & $23.33 \pm 0.38$ & $76.17 \pm 0.33$ & $77.10 \pm 0.50$ & $85.95 \pm 0.52$ & $62.09 \pm 0.35$ & $63.18 \pm 0.22$ & $66.14 \pm 0.46$ & $65.29 \pm 0.71$ \\
                    & 8     & $90.07 \pm 0.34$ & $61.85 \pm 0.22$ & $55.56 \pm 0.50$ & $71.40 \pm 2.80$ & $26.97 \pm 0.28$ & $76.72 \pm 0.31$ & $82.54 \pm 0.34$ & $86.52 \pm 0.45$ & $66.00 \pm 0.47$ & $65.25 \pm 0.48$ & $69.84 \pm 0.33$ & $68.43 \pm 0.59$ \\
                    & 16    & $90.79 \pm 0.15$ & $62.84 \pm 0.11$ & $61.74 \pm 0.61$ & $77.79 \pm 0.52$ & $31.75 \pm 0.46$ & $77.80 \pm 0.04$ & $86.91 \pm 0.71$ & $87.50 \pm 0.30$ & $71.28 \pm 0.20$ & $67.46 \pm 0.17$ & $72.20 \pm 0.03$ & $71.64 \pm 0.30$ \\ \midrule 
\multirow{5}{*}{Cross-Modal Linear Probe}  & 1     & $88.68 \pm 0.17$ & $60.19 \pm 0.14$ & \underline{$49.74 \pm 0.24$} & $59.54 \pm 5.28$ & $\mathbf{21.21 \pm 1.37}$ & $75.10 \pm 1.81$ & \underline{$80.35 \pm 0.22$} & $84.54 \pm 1.92$ & $58.68 \pm 0.17$ & $62.13 \pm 0.30$ & $65.24 \pm 0.36$ & $64.13 \pm 1.09$ \\
                    & 2     & $88.68 \pm 2.04$ & $60.56 \pm 0.10$ & $53.61 \pm 2.36$ & $65.23 \pm 2.42$ & \underline{$23.48 \pm 0.56$} & $77.27 \pm 0.07$ & $\mathbf{86.30 \pm 0.94}$ & $85.25 \pm 2.46$ & $61.75 \pm 0.29$ & $64.79 \pm 0.13$ & $69.53 \pm 0.74$ & $66.95 \pm 1.10$ \\
                    & 4     & $91.29 \pm 0.51$ & $61.48 \pm 0.15$ & \underline{$60.36 \pm 0.46$} & $72.72 \pm 2.00$ & \underline{$26.70 \pm 0.48$} & $77.27 \pm 0.66$ & $\mathbf{90.86 \pm 0.15}$ & $87.60 \pm 0.22$ & \underline{$65.88 \pm 0.06$} & $67.03 \pm 0.43$ & $72.24 \pm 0.35$ & $70.31 \pm 0.50$ \\
                    & 8     & $92.05 \pm 0.09$ & $62.44 \pm 0.08$ & $62.96 \pm 0.12$ & \underline{$79.21 \pm 2.13$} & $31.19 \pm 1.45$ & $78.38 \pm 0.19$ & $\mathbf{93.88 \pm 0.50}$ & $87.84 \pm 0.65$ & \underline{$69.76 \pm 0.63$} & \underline{$69.03 \pm 0.16$} & $75.86 \pm 0.37$ & $72.96 \pm 0.58$ \\
                    & 16    & $92.86 \pm 0.20$ & $64.51 \pm 0.05$ & $67.43 \pm 1.51$ & \underline{$84.91 \pm 0.27$} & $37.58 \pm 0.82$ & $78.57 \pm 0.54$ & $\mathbf{96.16 \pm 0.19}$ & $88.76 \pm 0.32$ & $75.49 \pm 0.36$ & $70.92 \pm 0.03$ & $78.47 \pm 0.12$ & \underline{$75.97 \pm 0.40$} \\ \midrule 
\multirow{5}{*}{Cross-Modal Wise-FT} & 1     & $88.61 \pm 0.15$ & $60.90 \pm 0.22$ & $48.17 \pm 0.17$ & $55.09 \pm 7.22$ & $20.62 \pm 0.44$ & $77.05 \pm 0.19$ & $77.18 \pm 1.70$ & \underline{$86.54 \pm 0.56$} & $\mathbf{59.10 \pm 0.40}$ & $62.47 \pm 0.32$ & $\underline{65.65 \pm 0.55}$ & $63.76 \pm 1.08$ \\
                    & 2     & $88.56 \pm 1.95$ & $61.77 \pm 0.16$ & $51.83 \pm 0.66$ & $64.33 \pm 3.76$ & $21.88 \pm 0.30$ & \underline{$77.62 \pm 0.21$} & $81.84 \pm 0.19$ & \underline{$87.01 \pm 0.12$} & $\mathbf{62.24 \pm 0.33}$ & $64.19 \pm 0.63$ & $69.11 \pm 0.92$ & $66.40 \pm 0.84$ \\
                    & 4     & $89.94 \pm 0.23$ & $62.45 \pm 0.13$ & $56.23 \pm 0.98$ & $72.22 \pm 2.18$ & $24.11 \pm 0.14$ & \underline{$78.25 \pm 0.09$} & $85.46 \pm 0.99$ & \underline{$87.99 \pm 0.22$} & $65.31 \pm 0.87$ & $65.61 \pm 0.57$ & $70.88 \pm 0.20$ & $68.95 \pm 0.60$ \\
                    & 8     & $91.36 \pm 0.27$ & $63.44 \pm 0.14$ & $60.15 \pm 2.36$ & $76.92 \pm 3.75$ & $28.59 \pm 2.21$ & $78.60 \pm 0.17$ & $90.72 \pm 0.97$ & \underline{$88.53 \pm 0.22$} & $68.57 \pm 1.41$ & $67.42 \pm 0.61$ & $74.83 \pm 1.18$ & $71.74 \pm 1.21$ \\
                    & 16    & $92.48 \pm 0.32$ & $65.15 \pm 0.05$ & $63.87 \pm 2.27$ & $79.96 \pm 1.76$ & $33.86 \pm 2.14$ & $78.94 \pm 0.38$ & $91.65 \pm 0.26$ & $\mathbf{89.38 \pm 0.21}$ & $73.64 \pm 0.66$ & $68.92 \pm 0.57$ & $77.12 \pm 0.56$ & $74.09 \pm 0.83$ \\ \midrule 
\multirow{5}{*}{Cross-Modal Adapter} & 1 & \underline{$89.03 \pm 0.36$} & \underline{$61.23 \pm 0.12$} & $47.24 \pm 0.91$ & \underline{$60.50 \pm 4.04$} & \underline{$21.04 \pm 1.30$} & $75.90 \pm 1.66$ & $\mathbf{80.63 \pm 0.28}$ & $85.62 \pm 0.71$ & \underline{$59.00 \pm 0.20$} & \underline{$62.86 \pm 0.24$} & $65.30 \pm 0.38$ & \underline{$64.40 \pm 0.93$} \\
                    & 2     & $89.36 \pm 1.20$ & \underline{$61.85 \pm 0.01$} & \underline{$54.51 \pm 1.55$} & $66.08 \pm 1.67$ & $\mathbf{23.58 \pm 0.62}$ & $77.53 \pm 0.20$ & \underline{$85.69 \pm 0.22$} & $86.89 \pm 0.23$ & \underline{$62.22 \pm 0.53$} & \underline{$65.46 \pm 0.26$} & \underline{$70.12 \pm 0.68$} & \underline{$67.57 \pm 0.65$} \\
                    & 4     & $\mathbf{91.33 \pm 0.23}$ & $\mathbf{62.98 \pm 0.10}$ & $60.03 \pm 0.53$ & \underline{$73.46 \pm 2.67$} & $\mathbf{27.55 \pm 0.47}$ & $77.92 \pm 0.63$ & \underline{$90.81 \pm 0.28$} & $87.76 \pm 0.12$ & $\mathbf{66.40 \pm 0.87}$ & \underline{$67.63 \pm 0.37$} & \underline{$72.67 \pm 0.04$} & $\mathbf{70.78 \pm 0.57}$ \\
                    & 8     & \underline{$92.08 \pm 0.02$} & $63.71 \pm 0.06$ & \underline{$64.11 \pm 0.91$} & $78.83 \pm 2.66$ & \underline{$32.75 \pm 0.14$} & \underline{$78.83 \pm 0.14$} & \underline{$93.57 \pm 0.19$} & $87.79 \pm 0.11$ & $\mathbf{70.29 \pm 0.45}$ & $68.61 \pm 0.52$ & \underline{$76.34 \pm 0.49$} & \underline{$73.35 \pm 0.52$} \\
                    & 16    & \underline{$92.98 \pm 0.14$} & $64.72 \pm 0.19$ & \underline{$67.51 \pm 1.32$} & $82.15 \pm 1.92$ & \underline{$38.80 \pm 1.06$} & $\mathbf{79.14 \pm 0.44}$ & $95.57 \pm 0.11$ & $88.64 \pm 0.16$ & $\mathbf{75.96 \pm 0.62}$ & $70.91 \pm 0.33$ & \underline{$78.91 \pm 0.14$} & $75.94 \pm 0.58$ \\ \midrule 
\multirow{5}{*}{Cross-Modal Partial Finetuning} & 1     & $\mathbf{89.10 \pm 0.36}$ & $\mathbf{61.55 \pm 0.45}$ & $\mathbf{49.92 \pm 0.76}$ & $\mathbf{61.84 \pm 5.16}$ & $20.56 \pm 0.21$ & \underline{$77.14 \pm 0.70$} & $76.25 \pm 0.42$ & $85.72 \pm 0.72$ & $58.96 \pm 0.15$ & $\mathbf{63.38 \pm 0.27}$ & $\mathbf{66.80 \pm 0.18}$ & $\mathbf{64.66 \pm 0.85}$ \\
                    & 2     & $\mathbf{89.97 \pm 0.28}$ & $\mathbf{62.64 \pm 0.12}$ & $\mathbf{55.18 \pm 1.77}$ & $\mathbf{68.48 \pm 1.75}$ & $22.65 \pm 0.72$ & $\mathbf{78.19 \pm 0.18}$ & $82.80 \pm 0.34$ & $\mathbf{87.24 \pm 0.99}$ & $61.19 \pm 0.36$ & $\mathbf{65.81 \pm 0.34}$ & $\mathbf{70.34 \pm 0.06}$ & $\mathbf{67.68 \pm 0.63}$ \\
                    & 4     & \underline{$91.30 \pm 0.75$} & \underline{$62.77 \pm 0.47$} & $\mathbf{60.68 \pm 0.36}$ & $\mathbf{75.21 \pm 2.10}$ & $25.58 \pm 0.61$ & $\mathbf{78.57 \pm 0.15}$ & $88.66 \pm 0.28$ & $87.86 \pm 0.73$ & $64.49 \pm 0.08$ & $\mathbf{67.76 \pm 0.51}$ & $\mathbf{73.61 \pm 0.09}$ & \underline{$70.59 \pm 0.56$} \\
                    & 8     & $\mathbf{92.20 \pm 0.19}$ & $\mathbf{64.23 \pm 0.11}$ & $\mathbf{64.72 \pm 0.54}$ & $\mathbf{81.33 \pm 1.61}$ & $\mathbf{33.87 \pm 0.70}$ & $\mathbf{78.92 \pm 0.21}$ & $93.50 \pm 0.24$ & $\mathbf{88.71 \pm 0.34}$ & $69.06 \pm 0.40$ & $\mathbf{69.64 \pm 0.08}$ & $\mathbf{77.50 \pm 1.04}$ & $\mathbf{73.97 \pm 0.50}$ \\
                    & 16    & $\mathbf{93.52 \pm 0.20}$ & $\mathbf{65.95 \pm 0.04}$ & $\mathbf{68.91 \pm 0.49}$ & $\mathbf{86.67 \pm 0.72}$ & $\mathbf{43.60 \pm 0.31}$ & $78.66 \pm 0.85$ & \underline{$95.72 \pm 0.22$} & $89.12 \pm 0.32$ & $75.45 \pm 0.49$ & $\mathbf{71.91 \pm 0.05}$ & $\mathbf{79.95 \pm 0.46}$ & $\mathbf{77.22 \pm 0.38}$ \\ \bottomrule
    \end{tabular}}
    \caption{\textbf{Per-dataset results on the ResNet-50 backbone.} We also include results from prior works for easier comparison. The zero-shot CLIP numbers differ from those reported in the original CLIP paper because we use one single prompt per dataset. We \textbf{bold} the best result for each shot and each dataset, and \underline{underline} the second best result. We see that cross-modal adaptation methods consistently produce the best performance across almost all dataset. The Tip-Adapter results are reproduced using only the few-shot validation set for hyperparameter searching and early stopping. }
    \label{tab:per_dataset}
\end{table*}

\begin{table*}[h!]
\centering
\resizebox{\linewidth}{!}{
\begin{tabular}{ccc|ccccc}
\toprule 
\multirow{2}{*}{Finetuning} & \multirow{2}{*}{ImageAug} & \multirow{2}{*}{TextAug} & \multicolumn{5}{c}{Number of shots} \\
\cmidrule(l){4-8}
& & & 1 & 2 & 4 & 8 & 16 \\
\midrule
\multirow{20}{*}{Linear} & CenterCrop (1 view) & \multirow{4}{*}{N/A (Uni-Modal Adaptation)} & $36.58_{(1.47)}$ & $48.85_{(1.43)}$ & $58.87_{(0.82)}$ & $66.46_{(0.74)}$ & $71.63_{(0.50)}$ \\
& +Flip (2 views) &  & $37.51_{(1.46)}$ & {\bf 49.43$_{(1.59)}$} & {\bf 59.37$_{(0.74)}$} & $66.65_{(0.64)}$ & $71.83_{(0.54)}$ \\
& +RandomCrop (2 views) & & $37.74_{(1.47)}$ & $49.21_{(1.46)}$ & $59.23_{(0.82)}$ & {\bf 66.70$_{(0.60)}$} & {\bf 71.94$_{(0.54)}$}  \\
& +RandomCrop (10 views) & & {\bf 37.76$_{(1.20)}$} & $49.25_{(1.14)}$ & $59.13_{(0.92)}$ & $66.52_{(0.59)}$ & $71.89_{(0.49)}$ \\ 
\cmidrule(l){2-8}
 & \multirow{4}{*}{CenterCrop (1 view)} & Class name & $61.78_{(1.17)}$ & $65.34_{(0.79)}$ & $68.98_{(0.67)}$ & $72.01_{(0.57)}$ & $74.91_{(0.59)}$ \\
& & {\tt a photo of a \{cls\}.} & $63.22_{(1.37)}$ & $66.18_{(0.74)}$ & $69.73_{(0.53)}$ & $72.51_{(0.71)}$ & $75.29_{(0.62)}$ \\
& & Hand Engineered & {\bf 63.66$_{(1.25)}$} & $66.67_{(0.91)}$ & {\bf 70.33$_{(0.53)}$} & $72.92_{(0.61)}$ & $75.54_{(0.53)}$ \\
& & Template Mining (21 views) & $63.50_{(1.33)}$ & {\bf 67.21$_{(0.80)}$} & $70.26_{(0.65)}$ & {\bf 73.07$_{(0.63)}$} & {\bf 75.73$_{(0.54)}$} \\
\cmidrule(l){2-8}
 & \multirow{4}{*}{+Flip (2 views)} & Class name & $61.84_{(0.79)}$ & $65.32_{(1.15)}$ & $69.25_{(0.52)}$ & $72.32_{(0.56)}$ & $75.27_{(0.49)}$ \\
& & {\tt a photo of a \{cls\}.} & $63.36_{(0.84)}$ & $66.42_{(1.20)}$ & $69.88_{(0.62)}$ & $72.73_{(0.71)}$ & $75.53_{(0.49)}$ \\
& & Hand Engineered & {\bf 64.13$_{(1.09)}$} & $66.95_{(1.10)}$ & $70.31_{(0.50)}$ & $72.96_{(0.58)}$ & {\bf 75.97$_{(0.40)}$} \\
& & Template Mining (21 views) & $63.88_{(1.21)}$ & {\bf 67.19$_{(0.97)}$} & {\bf 70.32$_{(0.70)}$} & {\bf 73.10$_{(0.57)}$} & $75.70_{(0.59)}$ \\
\cmidrule(l){2-8}
 & \multirow{4}{*}{+RandomCrop (2 views)} & Class name & $61.47_{(1.27)}$ & $65.09_{(1.20)}$ & $68.94_{(0.64)}$ & $72.06_{(0.76)}$ & $75.12_{(0.59)}$ \\
& & {\tt a photo of a \{cls\}.} & $63.32_{(1.14)}$ & $66.05_{(0.92)}$ & $69.93_{(0.63)}$ & $72.91_{(0.53)}$ & $75.67_{(0.50)}$  \\
& & Hand Engineered & {\bf 63.71$_{(1.50)}$} & $66.75_{(0.83)}$ & $70.19_{(0.51)}$ & $72.84_{(0.60)}$ & $75.83_{(0.59)}$ \\
& & Template Mining (21 views) & $63.68_{(1.75)}$ & {\bf 67.14$_{(0.80)}$} & {\bf 70.53$_{(0.53)}$} & {\bf 72.98$_{(0.67)}$} & {\bf 75.75$_{(0.49)}$} \\
\cmidrule(l){2-8}
 & \multirow{4}{*}{+RandomCrop (10 views)} & Class name & $61.52_{(1.18)}$ & $65.37_{(0.82)}$ & $68.85_{(0.77)}$ & $72.12_{(0.72)}$ & $75.02_{(0.63)}$ \\
& & {\tt a photo of a \{cls\}.} & $63.35_{(1.04)}$ & $66.45_{(0.73)}$ & $69.52_{(0.78)}$ & $72.69_{(0.55)}$ & $75.44_{(0.72)}$ \\
& & Hand Engineered & $63.85_{(1.35)}$ & $66.87_{(0.82)}$ & {\bf 70.19$_{(0.50)}$} & $72.98_{(0.59)}$ & $75.62_{(0.51)}$ \\
& & Template Mining (21 views) & {\bf 63.90$_{(1.35)}$} & {\bf 67.00$_{(0.86)}$} & $69.94_{(1.02)}$ & {\bf 73.04$_{(0.69)}$} & {\bf 75.75$_{(0.54)}$} \\
\hline
\multirow{20}{*}{Partial} & CenterCrop (1 view) & \multirow{4}{*}{N/A (Uni-Modal Adaptation)} & $29.93_{(2.37)}$ & $42.63_{(0.83)}$ & $54.27_{(1.06)}$ & $64.16_{(0.81)}$ & $71.62_{(0.56)}$ \\
& +Flip (2 views) & & {\bf 31.68$_{(1.19)}$} & $43.61_{(1.08)}$ & $55.15_{(0.77)}$ & $64.90_{(0.87)}$ & {\bf 72.19$_{(0.44)}$} \\
& +RandomCrop (2 views) & & $31.01_{(1.39)}$ & {\bf 43.78$_{(1.09)}$} & $55.16_{(0.79)}$ & {\bf 64.91$_{(0.93)}$} & $72.03_{(0.44)}$ \\
& +RandomCrop (10 views) & & $31.46_{(1.41)}$ & $43.76_{(1.07)}$ & {\bf 55.23$_{(0.79)}$} & $64.74_{(0.78)}$ & $72.15_{(0.41)}$ \\ 
\cmidrule(l){2-8}
 & \multirow{4}{*}{CenterCrop (1 view)} & Class name  & $62.50_{(1.34)}$ & $65.66_{(0.84)}$ & $69.33_{(0.86)}$ & $72.93_{(0.47)}$ & $76.21_{(0.41)}$ \\
& & {\tt a photo of a \{cls\}.} & $63.78_{(1.07)}$ & $66.79_{(0.68)}$ & $69.80_{(0.75)}$ & $73.40_{(0.43)}$ & $76.67_{(0.35)}$ \\
& & Hand Engineered & $64.27_{(0.96)}$ & $67.14_{(0.58)}$ & {\bf 70.26$_{(0.55)}$} & $73.53_{(0.51)}$ & $76.53_{(0.48)}$\\
& & Template Mining (21 views) & {\bf 64.57$_{(0.81)}$} & {\bf 67.21$_{(0.67)}$} & $70.24_{(0.89)}$ & {\bf 73.71$_{(0.58)}$} & {\bf 76.86$_{(0.32)}$} \\
\cmidrule(l){2-8}
 & \multirow{4}{*}{+Flip (2 views)} & Class name & $62.52_{(1.27)}$ & $66.02_{(0.86)}$ & $69.64_{(0.65)}$ & $73.30_{(0.59)}$ & $76.44_{(0.45)}$ \\
& & {\tt a photo of a \{cls\}.} & $64.13_{(0.97)}$ & $67.16_{(0.64)}$ & $69.97_{(1.22)}$ & $73.83_{(0.44)}$ & $77.03_{(0.39)}$\\
& & Hand Engineered & {\bf 64.66$_{(0.85)}$} & {\bf 67.68$_{(0.63)}$} & {\bf 70.59$_{(0.56)}$} & $73.79_{(0.50)}$ & {\bf 77.22$_{(0.38)}$} \\
& & Template Mining (21 views) & $64.59_{(1.02)}$ & $67.58_{(0.74)}$ & $70.58_{(0.82)}$ & {\bf 74.00$_{(0.49)}$} & $77.16_{(0.33)}$ \\
\cmidrule(l){2-8}
 & \multirow{4}{*}{+RandomCrop (2 views)} & Class name & $62.31_{(1.78)}$ & $65.77_{(0.77)}$ & $69.52_{(0.70)}$ & $73.21_{(0.49)}$ & $76.52_{(0.39)}$ \\
& & {\tt a photo of a \{cls\}.} & $63.72_{(1.09)}$ & $66.99_{(0.52)}$ & $69.89_{(1.14)}$ & $73.63_{(0.55)}$ & $76.94_{(0.37)}$\\
& & Hand Engineered & $63.64_{(1.54)}$ & $67.35_{(0.69)}$ & $70.50_{(0.69)}$ & {\bf 73.96$_{(0.48)}$} & $77.05_{(0.47)}$ \\
& & Template Mining (21 views) & {\bf 64.41$_{(1.18)}$} & {\bf 67.36$_{(0.75)}$} & {\bf 70.77$_{(0.61)}$} & $73.94_{(0.53)}$ & {\bf 77.19$_{(0.35)}$} \\
\cmidrule(l){2-8}
 & \multirow{4}{*}{+RandomCrop (10 views)} & Class name & $62.18_{(1.47)}$ & $66.01_{(0.64)}$ & $69.47_{(0.78)}$ & $73.27_{(0.46)}$ & $76.60_{(0.45)}$ \\
& & {\tt a photo of a \{cls\}.} & $64.00_{(1.12)}$ & $67.08_{(0.64)}$ & $70.22_{(0.64)}$ & $73.70_{(0.51)}$ & $76.96_{(0.41)}$ \\
& & Hand Engineered & $64.12_{(1.38)}$ & {\bf 67.63$_{(0.64)}$} & $70.58_{(0.59)}$ & $73.93_{(0.39)}$ & $77.13_{(0.38)}$ \\
& & Template Mining (21 views) & {\bf 64.57$_{(1.00)}$} & $67.37_{(0.62)}$ & {\bf 70.86$_{(0.54)}$} & {\bf 74.02$_{(0.41)}$} & {\bf 77.27$_{(0.38)}$} \\
    \bottomrule
    \end{tabular}
}
\caption{\small \textbf{Ablation for augmentation under vision-language adaptation.} Salient conclusions: (1) Uni-modal adaptation is much worse than cross-modal adaptation even when doing aggressive image augmentation to increase the number of views, e.g. 10 random crops. (2) Doing both image augmentation and text augmentation can improve the results, but text augmentation has a more profound impact whereas image augmentation saturates with a few views. (3) Simple template mining can be as competitive as manually selected templates (cf. \autoref{tab:180_templates}). Overall, we hope this preliminary investigation can encourage future work to explore more text augmentation strategies.}
\label{tab:augmentation_all}
\end{table*}

\begin{table*}[t]
\centering
\resizebox{\linewidth}{!}{
\begin{tabular}{cc|ccccc}
\toprule 
\multirow{2}{*}{Method} & \multirow{2}{*}{Initialization} & \multicolumn{5}{c}{Number of shots} \\
\cmidrule(l){3-7}
& & 1 & 2 & 4 & 8 & 16 \\
\midrule
\multirow{2}{*}{Linear Probing} & Random  & $36.58_{(1.47)}$ & $48.85_{(1.43)}$ & $58.87_{(0.82)}$ & $66.46_{(0.74)}$ & $71.63_{(0.50)}$\\
& Text & $58.32_{(0.71)}$	& $61.39_{(0.74)}$	& $65.25_{(0.61)}$	& $68.54_{(0.58)}$	& $71.90_{(0.33)}$		\\
\hline
\multirow{2}{*}{Cross-Modal Linear Probing} & Random & $48.37_{(1.58)}$ & $54.87_{(1.33)}$ & $61.98_{(0.84)}$ & $67.96_{(0.58)}$ & $72.32_{(0.50)}$\\
& Text & $63.66_{(1.25)}$ & $66.67_{(0.91)}$ & $70.33_{(0.53)}$ & $72.92_{(0.61)}$ & $75.54_{(0.53)}$	\\
\hline 
\multirow{2}{*}{Partial Finetuning} &  Random & $29.93_{(2.37)}$ & $42.63_{(0.83)}$& $54.27_{(1.06)}$& $64.16_{(0.81)}$& $71.62_{(0.56)}$\\
& Text & $60.79_{(1.53)}$ & $63.44_{(0.64)}$ & $66.51_{(0.60)}$ & $69.46_{(0.68)}$ & $72.67_{(0.54)}$\\
\hline
\multirow{2}{*}{Cross-Modal Partial Finetuning} & Random & $42.03_{(1.91)}$ & $50.85_{(1.20)}$ & $59.74_{(0.89)}$ & $66.98_{(0.90)}$ & $72.92_{(0.42)}$	\\
& Text & $64.27_{(0.96)}$ & $67.14_{(0.58)}$ & $70.26_{(0.55)}$ & $73.53_{(0.51)}$ & $76.53_{(0.48)}$	\\
\bottomrule
    \end{tabular}
}
\caption{\small \textbf{Ablation results for text-based vs random initialization for linear classifier weight.} We perform diligent analysis to confirm that initializing the linear classifier weights with text features is beneficial for the final performance. Still, cross-modal adaptation uniformly boosts the performance no matter the method or initialization. The text-based initialization is also more important for partial-finetuning than for linear probing, confirming the hypothesis~\cite{kumar2022fine} that a randomly initialized classifier will distort pre-trained features. Experiments in this table use center crop as image augmentation and Tip-Adapter's template as text augmentation for simplicity.}
\label{tab:init_results}
\end{table*}

\begin{table*}[t]
\centering
\resizebox{\linewidth}{!}{
\begin{tabular}{cc|ccccc}
\toprule 
\multirow{2}{*}{Image Encoder} & \multirow{2}{*}{Text Encoder} & \multicolumn{5}{c}{Number of shots} \\
\cmidrule(l){3-7}
& & 1 & 2 & 4 & 8 & 16 \\
\midrule
Frozen & Frozen & $63.66_{(1.25)}$ & $66.67_{(0.91)}$ & $70.33_{(0.53)}$ & $72.92_{(0.61)}$ & $75.54_{(0.53)}$\\
Finetune Attention Pooling Layer & Frozen & $64.13_{(1.29)}$ & $67.23_{(0.51)}$ & $70.44_{(0.55)}$ & $73.64_{(0.47)}$ & $76.65_{(0.44)}$\\
Frozen & Finetune Last Transformer Layer & $64.12_{(1.10)}$ & $67.41_{(0.79)}$ & $70.31_{(0.52)}$ & $72.12_{(0.38)}$ & $73.34_{(0.32)}$\\
Finetune Attention Pooling Layer & Finetune Last Transformer Layer & $64.09_{(1.28)}$ & $67.06_{(0.76)}$ & $70.38_{(0.57)}$ & $73.64_{(0.48)}$ & $76.68_{(0.39)}$\\
\bottomrule
    \end{tabular}
}
\caption{\small \textbf{Ablation results for partial-finetuning.} Partial finetuning of the last layer of image encoder is much more effective than finetuning the last layer of text encoder, suggesting that one may simply freeze the text encoder for few-shot vision-language adaptation. Experiments in this table use center crop as image augmentation and Tip-Adapter's template as text augmentation for simplicity.}
\label{tab:partial_results}
\end{table*}

\begin{table*}[t]
\centering
\resizebox{\linewidth}{!}{
\begin{tabular}{cc|ccccc}
\toprule 
\multirow{2}{*}{Backbone} & \multirow{2}{*}{Method} & \multicolumn{5}{c}{Number of shots} \\
\cmidrule(l){3-7}
& & 1 & 2 & 4 & 8 & 16 \\
\midrule
\multirow{7}{*}{ResNet50} & WiSE-FT & $59.09_{(0.69)}$ & $61.80_{(0.77)}$ & $65.29_{(0.71)}$ & $68.43_{(0.59)}$ & $71.64_{(0.30)}$\\
 & Cross-Modal WiSE-FT & $63.76_{(1.08)}$ & $66.40_{(0.84)}$ & $68.95_{(0.60)}$ & $71.74_{(1.21)}$ & $74.09_{(0.83)}$\\
 & Cross-Modal Prompting & $61.97_{(0.46)}$ & $64.91_{(0.48)}$ & $68.43_{(0.50)}$ & $71.39_{(0.59)}$ & $73.99_{0.54)}$ \\
 & Cross-Modal Adapter & $63.84_{(1.28)}$ & $67.11_{(0.96)}$ & $70.71_{(0.49)}$ & $73.32_{(0.67)}$ & $75.89_{(0.54)}$\\
 & Linear Probing & $36.58_{(1.47)}$ & $48.85_{(1.43)}$ & $58.87_{(0.82)}$ & $66.46_{(0.74)}$ & $71.63_{(0.50)}$ \\
 & Cross-Modal Linear Probing & $63.66_{(1.25)}$ & $66.67_{(0.91)}$ & $70.33_{(0.53)}$ & $72.92_{(0.61)}$ & $75.54_{(0.53)}$\\
 & Partial Finetuning  & $29.93_{(2.37)}$ & $42.63_{(0.83)}$ & $54.27_{(1.06)}$ & $64.16_{(0.81)}$ & $71.62_{(0.56)}$ \\
 & Cross-Modal Partial Finetuning & $64.27_{(0.96)}$ & $67.14_{(0.58)}$ & $70.26_{(0.55)}$ & $73.53_{(0.51)}$ & $76.53_{(0.48)}$\\
 \hline
\multirow{5}{*}{ViT-B/16} & WiSE-FT & $60.31_{(0.68)}$ & $62.27_{(0.72)}$ & $64.97_{(0.39)}$ & $67.03_{(0.44)}$ & $68.93_{(0.72)}$\\
& Cross-Modal WiSE-FT & $71.19_{(1.27)}$ & $73.45_{(0.79)}$ & $75.33_{(0.98)}$ & $77.91_{(0.85)}$ & $79.51_{(0.82)}$\\
& Linear Probing & $43.87_{(2.55)}$ & $56.84_{(1.45)}$ & $67.12_{(0.94)}$ & $73.77_{(0.69)}$ & $78.16_{(0.52)}$\\
 & Cross-Modal Linear Probing & $71.21_{(1.13)}$ & $73.70_{(1.03)}$ & $76.78_{(0.48)}$ & $78.89_{(0.37)}$ & $81.07_{(0.30)}$\\
 & Partial Finetuning & $35.44_{(3.49)}$ & $52.04_{(1.52)}$ & $65.50_{(0.99)}$ & $74.05_{(0.94)}$ & $79.58_{(0.53)}$\\
 & Cross-Modal Partial Finetuning & $70.70_{(1.21)}$ & $74.70_{(0.84)}$ & $77.76_{(0.50)}$ & $80.19_{(0.34)}$ & $82.52_{(0.41)}$ \\
 
\bottomrule
    \end{tabular}
}
\caption{\small \textbf{Complete results for all methods reported.} Experiments in this table use center crop as image augmentation and Tip-Adapter's template as text augmentation. Furthermore, we include ViT-B/16 results for completeness.}
\label{tab:complete_results}
\end{table*}

\begin{table*}[t]
\centering
\resizebox{\linewidth}{!}{
\begin{tabular}{cc|cccc}
\toprule 
\multirow{3}{*}{Dataset} & \multirow{3}{*}{Method} & \multicolumn{4}{c}{Number of Image Shots}  \\
\cmidrule(l){3-6}
& & 0 & 1 & 2 & 4 \\
\midrule
\multirow{5}{*}{ImageNet-ESC-19} & Image-Only Linear Probing & - & $68.00_{(4.17)}$	& $75.67_{(4.62)}$	& $83.05_{(2.52)}$ \\
 & Image-Audio Linear Probing & - & $69.33_{(3.97)}$	& $76.66_{(4.32)}$	& $83.22_{(3.77)}$ \\
  & Image-Text Linear Probing & - & $85.69_{(5.36)}$	& $86.94_{(2.41)}$	& $89.21_{(3.04)}$ \\
 & Image-Audio-Text Linear Probing & - & $82.34_{(2.66)}$	& $84.08_{(1.95)}$	& $87.33_{(1.68)}$ \\
 & Audio-initialized Classifier & $36.74_{(9.36)}$ & - & - & -\\
 & Text-initialized Classifier & $84.95_{(0.00)}$ & - & - & -\\
\hline
\multirow{5}{*}{ImageNet-ESC-27} & Image-Only Linear Probing & - & $60.13_{(3.97)}$ & $71.81_{(2.96)}$ & $79.01_{(2.50)}$ \\
 & Image-Audio Linear Probing & - & $60.87_{(4.41)}$	& $73.32_{(2.46)}$	& $78.94_{(2.66)}$ \\
 & Image-Text Linear Probing & - & $84.15_{(3.10)}$	& $85.17_{(2.48)}$	& $88.35_{(0.80)}$ \\
 & Image-Audio-Text Linear Probing & - & $75.96_{(2.77)}$	& $79.81_{(1.95)}$	& $83.41_{(1.19)}$ \\
 & Audio-initialized Classifier & $30.37_{(7.13)}$ & - & - & -\\
 & Text-initialized Classifier & $82.96_{(0.00)}$ & - & - & -\\
    \bottomrule
    \end{tabular}
}
\caption{\small \textbf{ImageNet-ESC image-classification results.} 
}
\label{tab:image_complete}
\end{table*}

\begin{table*}[t]
\centering
\resizebox{\linewidth}{!}{
\begin{tabular}{cc|cccc}
\toprule 
\multirow{3}{*}{Dataset} & \multirow{3}{*}{Method} & \multicolumn{4}{c}{Number of Audio Shots}  \\
\cmidrule(l){3-6}
& & 0 & 1 & 2 & 4 \\
\midrule
\multirow{5}{*}{ImageNet-ESC-19} & Audio-Only Linear Probing & - & $31.21_{(5.45)}$	& $41.11_{(5.12)}$	& $48.51_{(3.79)}$\\
 & Audio-Image Linear Probing & - & $35.74_{(4.85)}$	& $45.94_{(4.99)}$	& $51.59_{(3.40)}$ \\
  & Audio-Text Linear Probing & - & $38.74_{(5.51)}$	& $50.09_{(3.45)}$	& $53.90_{(1.96)}$ \\
 & Audio-Image-Text Linear Probing & - & $42.33_{(4.06)}$	& $49.32_{(4.67)}$	& $53.61_{(2.44)}$ \\
 & Image-initialized Classifier & $34.21_{(1.17)}$ & - & - & -\\
 & Text-initialized Classifier & $38.16_{(0.00)}$ & - & - & -\\
\hline
\multirow{5}{*}{ImageNet-ESC-27} & Audio-Only Linear Probing & - & $28.20_{(3.26)}$	& $39.00_{(3.42)}$	& $47.13_{(2.71)}$ \\
 & Audio-Image Linear Probing & - & $35.01_{(4.06)}$	& $43.51_{(3.47)}$	& $48.46_{(3.37)}$ \\
 & Audio-Text Linear Probing & - & $36.76_{(5.54)}$	& $45.69_{(4.04)}$	& $50.56_{(2.19)}$ \\
 & Audio-Image-Text Linear Probing & - & $36.06_{(5.36)}$	& $46.19_{(2.96)}$	& $50.79_{(2.49)}$ \\
 & Image-initialized Classifier & $29.00_{(0.84)}$ & - & - & -\\
 & Text-initialized Classifier & $31.02_{(0.00)}$ & - & - & -\\
    \bottomrule
    \end{tabular}
}
\caption{\small \textbf{ImageNet-ESC audio-classification results.} }
\label{tab:audio_complete}
\end{table*}

\begin{table*}[t]
\centering
\resizebox{\linewidth}{!}{
\begin{tabular}{cc|ccccc}
\toprule 
\multirow{2}{*}{Method} & \multirow{2}{*}{Template} & \multicolumn{5}{c}{Number of shots} \\
\cmidrule(l){3-7}
& & 1 & 2 & 4 & 8 & 16 \\
\midrule
ProDA~\cite{lu2022prompt} & 36 Learned Templates & \textbf{65.19} & \textbf{68.59} & \textbf{71.49} & 74.21 & 76.78\\
\hline
\multirow{4}{*}{Linear} & Class name & $62.34_{(0.88)}$ & $65.75_{(1.31)}$ & $69.95_{(0.53)}$ & $73.29_{(0.72)}$ & $76.66_{(0.30)}$ \\
 & {\tt a photo of a \{cls\}.}  & $63.87_{(0.88)}$ & $66.59_{(1.40)}$ & $70.71_{(0.61)}$ & $73.75_{(0.62)}$ & $76.85_{(0.38)}$  \\
 & HandEngineered~\cite{zhang2021tip} & $64.52_{(1.43)}$ & $67.31_{(1.26)}$ & $70.97_{(0.51)}$ & $73.77_{(0.84)}$ & $77.21_{(0.41)}$\\
& Template Mining (21 views)  & $64.37_{(1.38)}$ & $67.62_{(1.03)}$ & $71.00_{(0.70)}$ & $74.17_{(0.61)}$ & $77.15_{(0.47)}$ \\
\hline
\multirow{4}{*}{Partial} & Class name  & $62.58_{(1.87)}$ & $66.46_{(0.81)}$ & $70.29_{(0.61)}$ & $74.22_{(0.51)}$ & $77.73_{(0.57)}$\\
 & {\tt a photo of a \{cls\}.}  & $64.38_{(1.14)}$ & $67.48_{(0.67)}$ & $70.59_{(1.38)}$ & $74.68_{(0.45)}$ & $78.34_{(0.45)}$ \\
 & HandEngineered~\cite{zhang2021tip}  & $65.01_{(1.17)}$ & $68.05_{(0.64)}$ & $71.10_{(0.67)}$ & $74.83_{(0.50)}$ & {\bf 78.60$_{(0.40)}$} \\
 & Template Mining (21 views)  & $64.89_{(1.16)}$  & $68.03_{(0.74)}$  & $71.04_{(0.97)}$  & {\bf 74.90$_{(0.43)}$}  &  78.37$_{(0.40)}$ \\
    \bottomrule
    \end{tabular}
}
\caption{\small \textbf{Comparison to ProDA.} Since ProDA uses their own separate test split without releasing the code, it is not directly comparable to numbers reported in~\autoref{tab:comparison_to_sota}. Therefore, we reported results here with our best attempt to replicate their dataset split by using the official test splits of the datasets when available (Food101~\cite{bossard2014food} and DTD~\cite{cimpoi2014describing}). Note that ProDA reported results using 36 learned prompts, whereas our template mining only uses 21 templates searched on few-shot validation set without any learning. Since we do not know whether ProDA uses augmentation, we report center crop results in this table. Still, our approach is generally more performant than ProDA and we do not require deep finetuning which takes 100x training time.}
\label{tab:comparison_to_proda}
\end{table*}

\begin{table*}[t]
\centering
\resizebox{\linewidth}{!}{
\begin{tabular}{c|c|c}
\toprule 
\multicolumn{3}{c}{{\bf 180 Templates ($*$ indicates not in CoOp codebase)}} \\
{\tt \{cls\}}$^{\boldsymbol{*}}$	&	{\tt a tattoo of the \{cls\}.}	&	{\tt a video of the person \{cls\}.} \\
{\tt a photo of a \{cls\}.}$^{\boldsymbol{*}}$	&	{\tt a photo of a person during \{cls\}.}	&	{\tt a example of a person \{cls\}.} \\
{\tt a picture of this \{cls\}.}$^{\boldsymbol{*}}$	&	{\tt a photo of a clean \{cls\}.}	&	{\tt a photo of a small \{cls\}.} \\
{\tt a photo of my \{cls\}.}$^{\boldsymbol{*}}$	&	{\tt a photo of a \{cls\} texture.}	&	{\tt a photo of the small \{cls\}.} \\
{\tt that is a \{cls\} photo.}$^{\boldsymbol{*}}$	&	{\tt a bad photo of a \{cls\}.}	&	{\tt the \{cls\} in a video game.} \\
{\tt a picture of a \{cls\}.}$^{\boldsymbol{*}}$	&	{\tt a video of the person during \{cls\}.}	&	{\tt a demonstration of a person \{cls\}.} \\
{\tt a \{cls\} photo.}$^{\boldsymbol{*}}$	&	{\tt a drawing of the \{cls\}.}	&	{\tt a photo of one \{cls\}.} \\
{\tt this is a \{cls\} photo.}$^{\boldsymbol{*}}$	&	{\tt a close-up photo of the \{cls\}.}	&	{\tt a video of a person using \{cls\}.} \\
{\tt a photo of these \{cls\}.}$^{\boldsymbol{*}}$	&	{\tt a video of a person \{cls\}.}	&	{\tt a blurry photo of a \{cls\}.} \\
{\tt a picture of my \{cls\}.}$^{\boldsymbol{*}}$	&	{\tt a good photo of a \{cls\}.}	&	{\tt a photo of a person practicing \{cls\}.} \\
{\tt a \{cls\} picture.}$^{\boldsymbol{*}}$	&	{\tt a photo of a \{cls\} thing.}	&	{\tt a photo of a \{cls\}, a type of flower.} \\
{\tt that is a \{cls\} picture.}$^{\boldsymbol{*}}$	&	{\tt a demonstration of the person practicing \{cls\}.}	&	{\tt a painting of a \{cls\}.} \\
{\tt a picture of those \{cls\}.}$^{\boldsymbol{*}}$	&	{\tt itap of a \{cls\}.}	&	{\tt a example of the person \{cls\}.} \\
{\tt this is a \{cls\} picture.}$^{\boldsymbol{*}}$	&	{\tt a photo of a \{cls\} pattern.}	&	{\tt a example of the person performing \{cls\}.} \\
{\tt that is a photo of a \{cls\}.}$^{\boldsymbol{*}}$	&	{\tt itap of the \{cls\}.}	&	{\tt a rendition of the \{cls\}.} \\
{\tt a photo of your \{cls\}.}$^{\boldsymbol{*}}$	&	{\tt a demonstration of a person using \{cls\}.}	&	{\tt a cropped photo of a \{cls\}.} \\
{\tt a picture of some \{cls\}.}$^{\boldsymbol{*}}$	&	{\tt a cropped photo of the \{cls\}.}	&	{\tt the origami \{cls\}.} \\
{\tt a photo of those \{cls\}.}$^{\boldsymbol{*}}$	&	{\tt a example of the person practicing \{cls\}.}	&	{\tt a photo of the person \{cls\}.} \\
{\tt a picture of these \{cls\}.}$^{\boldsymbol{*}}$	&	{\tt a bright photo of a \{cls\}.}	&	{\tt a example of the person doing \{cls\}.} \\
{\tt \{cls\}, a picture.}$^{\boldsymbol{*}}$	&	{\tt a photo of the hard to see \{cls\}.}	&	{\tt a photo of the large \{cls\}.} \\
{\tt a photo of an \{cls\}.}$^{\boldsymbol{*}}$	&	{\tt a photo of a person using \{cls\}.}	&	{\tt a example of a person doing \{cls\}.} \\
{\tt a picture of the \{cls\}.}$^{\boldsymbol{*}}$	&	{\tt a rendition of a \{cls\}.}	&	{\tt a video of a person doing \{cls\}.} \\
{\tt \{cls\}, a photo.}$^{\boldsymbol{*}}$	&	{\tt a demonstration of a person during \{cls\}.}	&	{\tt a sketch of the \{cls\}.} \\
{\tt a photo of this \{cls\}.}$^{\boldsymbol{*}}$	&	{\tt graffiti of the \{cls\}.}	&	{\tt a photo of a nice \{cls\}.} \\
{\tt a photo of the \{cls\}.}$^{\boldsymbol{*}}$	&	{\tt a toy \{cls\}.}	&	{\tt a good photo of the \{cls\}.} \\
{\tt this is a photo of a \{cls\}.}$^{\boldsymbol{*}}$	&	{\tt a jpeg corrupted photo of the \{cls\}.}	&	{\tt a photo of a person performing \{cls\}.} \\
{\tt a picture of your \{cls\}.}$^{\boldsymbol{*}}$	&	{\tt a photo of the weird \{cls\}.}	&	{\tt a pixelated photo of the \{cls\}.} \\
{\tt a photo of a \{cls\}.}$^{\boldsymbol{*}}$	&	{\tt a photo of a cool \{cls\}.}	&	{\tt a photo of the dirty \{cls\}.} \\
{\tt a picture of that \{cls\}.}$^{\boldsymbol{*}}$	&	{\tt a video of the person practicing \{cls\}.}	&	{\tt a photo of my new \{cls\}.} \\
{\tt a photo of some \{cls\}.}$^{\boldsymbol{*}}$	&	{\tt the plushie \{cls\}.}	&	{\tt a sculpture of the \{cls\}.} \\
{\tt a photo of my \{cls\}.}$^{\boldsymbol{*}}$	&	{\tt a low resolution photo of a \{cls\}.}	&	{\tt a photo of the person doing \{cls\}.} \\
{\tt a photo of the \{cls\}.}$^{\boldsymbol{*}}$	&	{\tt a photo of the person performing \{cls\}.}	&	{\tt a photo of a \{cls\}, a type of pet.} \\
{\tt a photo of that \{cls\}.}$^{\boldsymbol{*}}$	&	{\tt the cartoon \{cls\}.}	&	{\tt a centered satellite photo of the \{cls\}.} \\
{\tt a picture of an \{cls\}.}$^{\boldsymbol{*}}$	&	{\tt a video of a person practicing \{cls\}.}	&	{\tt a photo of the \{cls\} texture.} \\
{\tt a photo of the \{cls\}, a type of aircraft.}	&	{\tt a photo of a \{cls\}, a type of aircraft.}	&	{\tt a photo of a hard to see \{cls\}.} \\
{\tt a bad photo of the \{cls\}.}	&	{\tt a photo of the person using \{cls\}.}	&	{\tt a black and white photo of a \{cls\}.} \\
{\tt a photo of my dirty \{cls\}.}	&	{\tt a centered satellite photo of \{cls\}.}	&	{\tt itap of my \{cls\}.} \\
{\tt a example of a person during \{cls\}.}	&	{\tt a example of a person performing \{cls\}.}	&	{\tt a video of the person doing \{cls\}.} \\
{\tt a demonstration of the person doing \{cls\}.}	&	{\tt a \{cls\} in a video game.}	&	{\tt a demonstration of the person performing \{cls\}.} \\
{\tt a demonstration of a person performing \{cls\}.}	&	{\tt i love my \{cls\}!}	&	{\tt art of a \{cls\}.} \\
{\tt a photo of the person practicing \{cls\}.}	&	{\tt a example of a person using \{cls\}.}	&	{\tt a black and white photo of the \{cls\}.} \\
{\tt a photo of a large \{cls\}.}	&	{\tt a example of the person using \{cls\}.}	&	{\tt a photo of the clean \{cls\}.} \\
{\tt a photo of a weird \{cls\}.}	&	{\tt a jpeg corrupted photo of a \{cls\}.}	&	{\tt a photo of the nice \{cls\}.} \\
{\tt a photo of a person \{cls\}.}	&	{\tt a blurry photo of the \{cls\}.}	&	{\tt a doodle of the \{cls\}.} \\
{\tt a video of a person during \{cls\}.}	&	{\tt a painting of the \{cls\}.}	&	{\tt a close-up photo of a \{cls\}.} \\
{\tt a photo of the \{cls\} thing.}	&	{\tt a sculpture of a \{cls\}.}	&	{\tt a low resolution photo of the \{cls\}.} \\
{\tt the embroidered \{cls\}.}	&	{\tt a demonstration of the person using \{cls\}.}	&	{\tt a dark photo of a \{cls\}.} \\
{\tt a photo of a \{cls\} object.}	&	{\tt a sketch of a \{cls\}.}	&	{\tt a video of the person performing \{cls\}.} \\
{\tt a dark photo of the \{cls\}.}	&	{\tt a drawing of a \{cls\}.}	&	{\tt a photo of a dirty \{cls\}.} \\
{\tt a photo of \{cls\}, a type of food.}	&	{\tt a photo of the \{cls\} pattern.}	&	{\tt a cartoon \{cls\}.} \\
{\tt a example of the person during \{cls\}.}	&	{\tt a photo of the cool \{cls\}.}	&	{\tt the plastic \{cls\}.} \\
{\tt a video of a person performing \{cls\}.}	&	{\tt a photo of the \{cls\} object.}	&	{\tt a photo of my clean \{cls\}.} \\
{\tt a photo of many \{cls\}.}	&	{\tt a video of the person using \{cls\}.}	&	{\tt a photo of my old \{cls\}.} \\
{\tt a photo of a person doing \{cls\}.}	&	{\tt a demonstration of the person during \{cls\}.}	&	{\tt a pixelated photo of a \{cls\}.} \\
{\tt a plushie \{cls\}.}	&	{\tt a centered satellite photo of a \{cls\}.}	&	{\tt a demonstration of the person \{cls\}.} \\
{\tt art of the \{cls\}.}	&	{\tt a tattoo of a \{cls\}.}	&	{\tt a doodle of a \{cls\}.} \\
{\tt a photo of the person during \{cls\}.}	&	{\tt graffiti of a \{cls\}.}	&	{\tt the toy \{cls\}.} \\
{\tt a bright photo of the \{cls\}.}	&	{\tt a demonstration of a person practicing \{cls\}.}	&	{\tt a plastic \{cls\}.} \\
{\tt a rendering of a \{cls\}.}	&	{\tt a embroidered \{cls\}.}	&	{\tt a rendering of the \{cls\}.} \\
{\tt a origami \{cls\}.}	&	{\tt a example of a person practicing \{cls\}.}	&	{\tt a demonstration of a person doing \{cls\}.} \\
    \bottomrule
    \end{tabular}
}
\caption{\small \textbf{Templates used during template mining.} Most of the templates we use come from the original CoOp codebase~\cite{zhou2022coop}. In addition, we add 31 random templates by paraphrasing~\cite{jiang2020can} the standard template {\tt a photo of a \{cls\}}. We encourage future work to try out more sophisticated techniques to generate templates, e.g. through automated prompting~\cite{zhou2022coop} or with the help of language models~\cite{jiang2020can}.}
\label{tab:180_templates}
\end{table*}

\end{document}